\documentclass{ieeeaccess}
\usepackage{cite}
\usepackage{amsmath,amssymb,amsfonts}
\usepackage{algorithmic}
\usepackage{graphicx}
\usepackage{textcomp}

\usepackage{soul}   
\usepackage{tablefootnote}
\usepackage{bm}
\makeatletter
\AtBeginDocument{\DeclareMathVersion{bold}
\SetSymbolFont{operators}{bold}{T1}{times}{b}{n}
\SetSymbolFont{NewLetters}{bold}{T1}{times}{b}{it}
\SetMathAlphabet{\mathrm}{bold}{T1}{times}{b}{n}
\SetMathAlphabet{\mathit}{bold}{T1}{times}{b}{it}
\SetMathAlphabet{\mathbf}{bold}{T1}{times}{b}{n}
\SetMathAlphabet{\mathtt}{bold}{OT1}{pcr}{b}{n}
\SetSymbolFont{symbols}{bold}{OMS}{cmsy}{b}{n}
\renewcommand\boldmath{\@nomath\boldmath\mathversion{bold}}}
\makeatother

\def\BibTeX{{\rm B\kern-.05em{\sc i\kern-.025em b}\kern-.08em
    T\kern-.1667em\lower.7ex\hbox{E}\kern-.125emX}}

\usepackage{scalerel}
\usepackage{tikz}
\usetikzlibrary{svg.path}
\NewSpotColorSpace{PANTONE}
  \AddSpotColor{PANTONE} {PANTONE3015C} {PANTONE\SpotSpace 3015\SpotSpace C} {1 0.3 0 0.2}
  \SetPageColorSpace{PANTONE}
\definecolor{orcidlogocol}{HTML}{A6CE39}
\tikzset{
  orcidlogo/.pic={
    \fill[orcidlogocol] svg{M256,128c0,70.7-57.3,128-128,128C57.3,256,0,198.7,0,128C0,57.3,57.3,0,128,0C198.7,0,256,57.3,256,128z};
    \fill[white] svg{M86.3,186.2H70.9V79.1h15.4v48.4V186.2z}
                 svg{M108.9,79.1h41.6c39.6,0,57,28.3,57,53.6c0,27.5-21.5,53.6-56.8,53.6h-41.8V79.1z M124.3,172.4h24.5c34.9,0,42.9-26.5,42.9-39.7c0-21.5-13.7-39.7-43.7-39.7h-23.7V172.4z}
                 svg{M88.7,56.8c0,5.5-4.5,10.1-10.1,10.1c-5.6,0-10.1-4.6-10.1-10.1c0-5.6,4.5-10.1,10.1-10.1C84.2,46.7,88.7,51.3,88.7,56.8z};
  }
}

\newcommand\orcidicon[1]{\href{https://orcid.org/#1}{\mbox{\scalerel*{
\begin{tikzpicture}[yscale=-1,transform shape]
\pic{orcidlogo};
\end{tikzpicture}
}{|}}}}

\usepackage{hyperref} 

\hypersetup{
  colorlinks=false,
  linkbordercolor=white,
 urlbordercolor=white,
pdfborder={0 0 0}
}

\begin{document}
\history{Date of publication xxxx 00, 0000, date of current version xxxx 00, 0000.}
\doi{10.1109/ACCESS.2023.1120000}

\title{Achieving Peak Performance for Large Language Models: A Systematic Review}
\author{\uppercase{Zhyar Rzgar K Rostam  \textsuperscript{\orcidicon{0000-0001-9906-6883}}}\authorrefmark{1}\IEEEmembership{Member, IEEE}, \uppercase{Sándor Szénási  \textsuperscript{\orcidicon{0000-0002-7292-0717}}}\authorrefmark{2,3} \IEEEmembership{Member, IEEE}, and \uppercase {Gábor Kertész  \textsuperscript{\orcidicon{0000-0002-8845-8301}}}\authorrefmark{2,4} \IEEEmembership{Senior Member, IEEE}}

\address[1]{Doctoral School of Applied Informatics and Applied Mathematics, Óbuda University, Budapest, Hungary}
\address[2]{John von Neumann Faculty of Informatics, Óbuda University, Budapest, Hungary}
\address[3]{Faculty of Economics and Informatics, J. Selye University, Komarno, Slovakia}
\address[4]{Laboratory of Parallel and Distributed Systems, Institute for Computer Science and Control (SZTAKI), Hungarian Research Network (HUN-REN), Budapest, Hungary}

\markboth
{Author \headeretal: Preparation of Papers for IEEE TRANSACTIONS and JOURNALS}
{Author \headeretal: Preparation of Papers for IEEE TRANSACTIONS and JOURNALS}

\corresp{Corresponding author: Zhyar Rzgar K Rostam (e-mail: zhyar.rostam@stud.uni-obuda.hu).}

\begin{abstract}
In recent years, large language models (LLMs) have achieved remarkable success in natural language processing (NLP). LLMs require an extreme amount of parameters to attain high performance. As models grow into the trillion-parameter range, computational and memory costs increase significantly. This makes it difficult for many researchers to access the resources needed to train or apply these models. Optimizing LLM performance involves two main approaches: fine-tuning pre-trained models for specific tasks to achieve state-of-the-art performance, and reducing costs or improving training time while maintaining similar performance. This paper presents a systematic literature review (SLR) following the Preferred Reporting Items for Systematic Reviews and Meta-Analyses (PRISMA) statement. We reviewed 65 publications out of 983 from 2017 to December 2023, retrieved from 5 databases. The study presents methods to optimize and accelerate LLMs while achieving cutting-edge results without sacrificing accuracy. We begin with an overview of the development of language modeling, followed by a detailed explanation of commonly used frameworks and libraries, and a taxonomy for improving and speeding up LLMs based on three classes: LLM training, LLM inference, and system serving. We then delve into recent optimization and acceleration strategies such as training optimization, hardware optimization, scalability and reliability, accompanied by the taxonomy and categorization of these strategies. Finally, we provide an in-depth comparison of each class and strategy, with two case studies on optimizing model training and enhancing inference efficiency. These case studies showcase practical approaches to address LLM resource limitations while maintaining performance.

\end{abstract}

\begin{keywords}
Distributed training, GPU acceleration, Large Language Model, LLM, LLM Acceleration, LLM frameworks, LLM Optimization.
\end{keywords}

\titlepgskip=-21pt

\maketitle

\section{Introduction}
\label{sec:introduction}
In recent years, dense deep learning models have seen an extraordinary growth in the number of parameters \cite{b1, b2, b3}. Transformer as an effective deep learning architecture has been widely used over the recent years, and transformer-based models have achieved notable success and recognition in various fields including language modeling compared to the existing models \cite{b4, b5, b6, b7, b8, b9, b10, b11, b12, b13_u}.

To achieve significant accuracy in deep learning, large models with billions to trillions of parameters are essential. Therefore, deep learning models continue to grow in complexity with an array of large-scale models ranging from Bidirectional Encoder Representations from Transformers ($BERT_{large}$, 340 million parameters) \cite{b8}, Generative Pre-trained Transformer-3 (GPT-3, 175 billion parameters) \cite{b13}, to General Language Model (GLM-3, 1.75 trillion parameters) \cite{b14}. With models now reaching trillions of parameters, even the most powerful GPUs are struggling to keep up \cite{b1}. This resource-intensive requirement is making it difficult for many researchers to access the computational resources they need to train these models \cite{b1, b4, b15}. Also, handling, managing, and fitting these models into device memory is a daunting challenge due to memory limitations, and this tremendous size of data brings complexity, and requires high-end computing resources with significant memory requirements to process \cite{b5, b16, b17, b18}. Training large-scale models effectively require significant adjustments \cite{b19, b20, b21, b22, b23}, especially in terms of increasing training throughput and loading these kinds of large models into GPU memory \cite{b17}.

As a result, developing frameworks, libraries and proposing new techniques to overcome the mentioned challenges has become an essential task. There are many studies that have worked on possibilities for optimization and acceleration with large models and using various techniques to achieve state-of-the-art (SOTA) results without sacrificing accuracy. These remarkable advancements in the field of language models (LMs) required a systematic review of recent LM optimization and acceleration techniques. To address these challenges and guide future research, this SLR paper aims to:
\begin{itemize}
    \item Analyze recent optimization and acceleration techniques for LLMs.
    \item Identify challenges associated with training, inference, and system serving for LLMs (billions/trillions of parameters).
    \item Develop a structured taxonomy to categorize LLM optimization techniques.
    \item Review and evaluate recent libraries and frameworks designed for LLM optimization.
    \item Identify promising areas for future research in LLM development, focusing on efficiency, scalability, and flexibility.
\end{itemize}

In this SLR we are making the following contributions:
\begin{itemize}
    \item Comprehensive overview: We offer a comprehensive overview of the development of language modeling (Section \ref{sec:language_modeling_development}), detailing commonly used frameworks and libraries (Section \ref{sec:frameworks_and_libraries}), and recently used techniques and strategies (Sections \ref{sec:training_optimization}, \ref{sec:hardware_optimization}, \ref{sec:scalability_and_reliability}). This serves as a valuable resource for understanding the current landscape of LLM optimization.
    \item Taxonomy of optimization strategies: We categorize optimization strategies into three classes: training optimization, hardware optimization, and scalability and reliability. This taxonomy helps clarify the various approaches and their specific applications (presented in Fig \ref{fig4}, Sections \ref{sec:training_optimization}, \ref{sec:hardware_optimization}, \ref{sec:scalability_and_reliability}).
    \item Detailed analysis of techniques: Our analysis explores recent optimization and acceleration strategies, we provide two comparative analyses regarding performance, cost, and scalability for reviewed strategies (presented in Tables \ref{rosta.t6} and \ref{rosta.t7}) and their core categories: training optimization, hardware optimization, and scalability and reliability (presented in Table \ref{rosta.t5}). In the latter analysis, we also consider the focus of classes.
    \item Case studies: We include two in-depth case studies that demonstrate practical approaches to optimizing model training and enhancing inference efficiency. These case studies highlight how resource limitations can be addressed while maintaining performance (Sections \ref{subsec:optimizing_model_training_with_sparsegpt}, \ref{subsec:enhancing_inference_efficiency_with_QMoE}).
    \item Future direction: We explore a range of promising future directions for LLM development. These areas, detailed in specific sections, focus on enhancing efficiency, scalability, and flexibility for LLMs (Section \ref{conclusion_and_future_directions}).
\end{itemize}
This review paper is organized as follows: an overview of language modeling development (Section \ref{sec:language_modeling_development}), followed by an in-depth explanation of the most commonly utilized frameworks and libraries specifically designed for optimizing and accelerating large language models (LLMs) (Section \ref{sec:frameworks_and_libraries}, Tables \ref{rosta.t3} and \ref{rosta.t4}), accompanied by taxonomy and categorization. Additionally, it delves into recent optimization and acceleration strategies employed within LLMs, including the taxonomy and categorization of these strategies (presented in Fig. \ref{fig1}) (Section \ref{sec:training_optimization}, \ref{sec:hardware_optimization}, \ref{sec:scalability_and_reliability}), Table \ref{rosta.t8} summarizes the reviewed papers, excluding those already covered in Tables \ref{rosta.t3} and \ref{rosta.t4} or the main text. Moreover, we present an individual comparison in terms of performance, cost, and scalability for reviewed strategies discussed in Tables \ref{rosta.t6}, and \ref{rosta.t7}, and the classes (training optimization, hardware optimization, scalability and reliability) presented in Table \ref{rosta.t5}. In addition to the mentioned factors, we consider the classes' focus in this comparison. Finally, we illustrate these concepts with two real-world examples: optimizing model training and improving inference efficiency through case studies (Section \ref{sec:case_studies}). 

\Figure[t!](topskip=0pt, botskip=0pt, midskip=0pt)[width=0.9\linewidth]{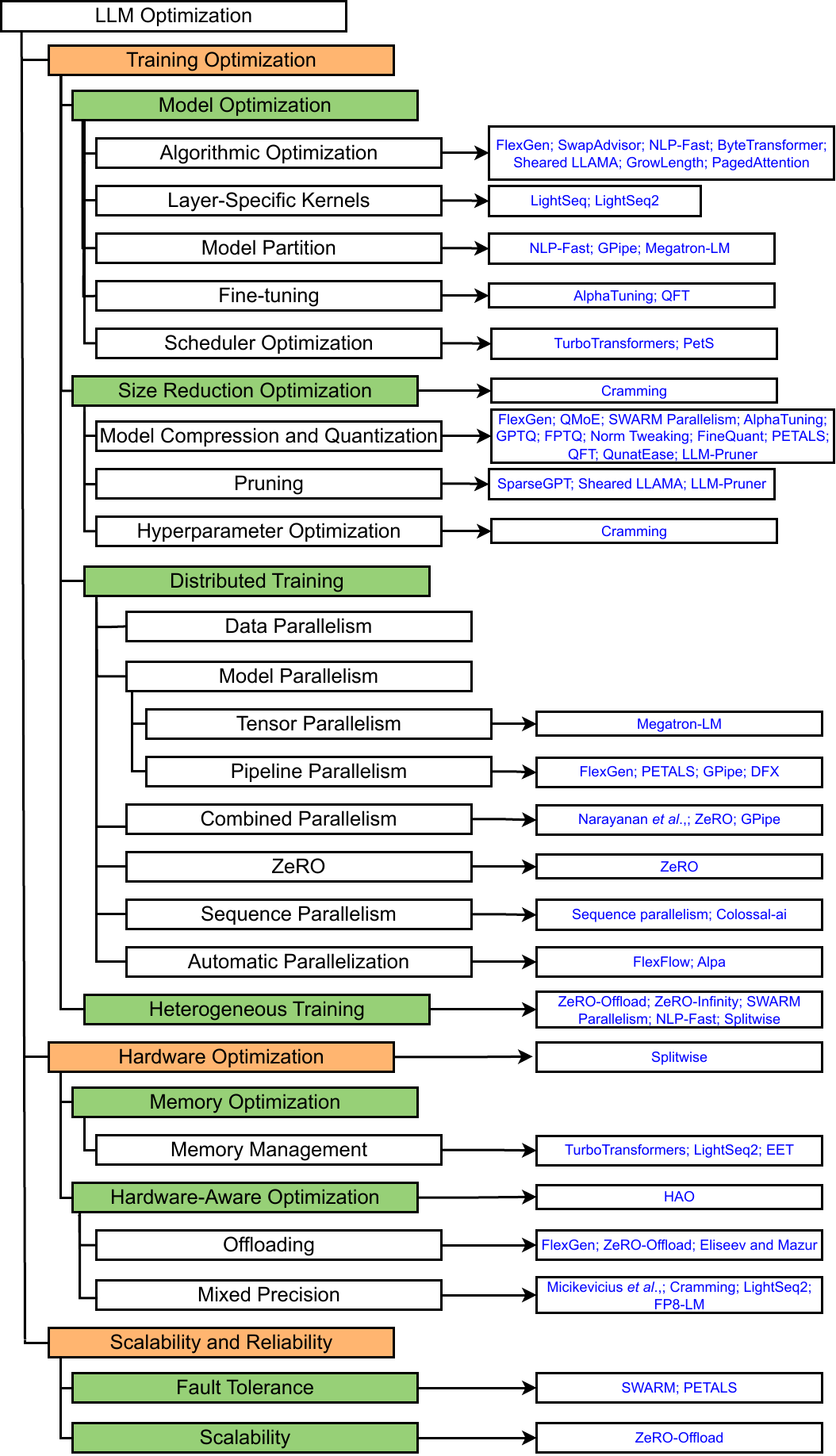}
{\textbf{LLM optimization techniques and taxonomy}\label{fig1}}

\subsection{Related works}
In this section, we will present the related studies that investigate optimization and acceleration with dense deep learning models and LLMs. Jahan \textit{et al}., in \cite{b24} present a systematic literature review (SLR) by comparing 31 language models inspired by BERT, published between 2018 and 2020, to help researchers choose the best model based on their requirements. By analyzing each model's performance against RoBERTa, the study identified seven models that performed better, and the rest of the studies investigated with different parameter settings. The outperforming models varied in dataset size, suggesting that both large and small datasets can be effective depending on the model's architecture. Ultimately, this research provides valuable insights for researchers seeking the optimal language model for their specific tasks. Yu \textit{et al} \cite{b25} conduct a survey that explores the growing challenges and opportunities for optimizing large-scale deep learning systems. By highlighting recent advances in optimization techniques, it proposes a new way to categorize and explain the different computing approaches used. Zhao \textit{et al} \cite{b26} carry out a survey that focuses on the recent advancements in LLMs. The study concentrates on four major dimensions of LLMs: pre-training, adaptation tuning, utilization, and capacity evaluation. The survey emphasizes the techniques or discoveries essential for LLMs' success. Additionally, it provides an overview of available development resources and offers valuable guidelines for successful LLM implementation, drawing from the latest research. Bai \textit{et al}., in \cite{b27} provide a systematic survey that provides an overview of LLM resource efficiency. It focuses on LLM significant resource consumption in computational, memory, energy, and financial aspects. It categorizes techniques aimed at improving LLMs' resource efficiency. Standardized evaluation metrics and datasets are also proposed to facilitate fair comparisons. The survey offers insights into current advancements and guides future developments toward more sustainable and efficient LLMs. Wang \textit{et al} \cite{b28} explore new methods to achieve comparable accuracy while reducing training costs. They highlight optimized algorithms for quicker learning, distributed architectures leveraging widespread computing resources, and hardware acceleration with communication optimization for collaborative training. While challenges remain, these advancements pave the way for more affordable and accessible AI in the future. Min \textit{et al} \cite{b29} present a survey that explored recent studies for using powerful pre-trained language models (PLMs) in natural language processing (NLP) tasks through the analysis of three popular approaches. The first approach trains on a massive dataset for general language understanding, then specializes it for a specific task with focused training. The second approach prompts the PLM to treat the desired task as similar to its pre-training tasks, allowing for efficient ``few-shot'' learning with just a few examples. The third approach modifies NLP tasks as text generation to maximize the utilization of knowledge embedded within a generative language model. Qiu \textit{et al} \cite{b30} provide a comprehensive overview of pre-trained models (PTMs), from fundamental knowledge and model architectures to diverse pre-training tasks, extensions, and real-world applications. It proposes a clear taxonomy for easy navigation and provides abundant resources like code, tools, corpora, and reading lists. 
Recognizing current limitations, the survey also presents a discussion on promising future directions to shape the NLP landscape. This survey \cite{b31} explores techniques for building efficient LLMs. It categorizes approaches into three groups: Model-centric, Data-centric, and LLM frameworks. In the model-centric method focuses on optimizing the LLMs through techniques including compression, efficient training, and specialized architectures. Data-centric focusing on improving data quality and using prompts to guide the model efficiently. LLM frameworks create specialized software to handle LLMs, the survey aims to provide a comprehensive understanding of how to make LLMs more efficient and accessible.

In this SLR, we examine research published between 2017 and December 2023, filling a gap in existing surveys by specifically focusing on optimizing and speeding up LLMs. Following the PRISMA approach, we reviewed 65 articles. The study starts with an overview of language modeling's development and then dives deep into the most popular frameworks and libraries for optimizing and accelerating LLMs. It organizes these models with a clear taxonomy and categorizes them effectively. The research also investigates recent approaches for optimizing and accelerating LLMs, offering a classification system along with a summary and comparison of the reviewed papers containing the latest optimization techniques. Moreover, resource limitations and the impact of various optimization techniques in LLMs were addressed through two in-depth case studies. These studies delve into practical approaches for optimizing training and enhancing inference efficiency, demonstrating how these techniques can be applied effectively without excessive resources.

\subsection{Research Methodology}
\label{subsec:research_methodology}

In this study, we have followed the PRISMA statement to ensure a systematic and transparent methodology. PRISMA provides a comprehensive set of guidelines for conducting systematic reviews. Our approach included a detailed search strategy across multiple databases, explicit inclusion and exclusion criteria, and a thorough study selection process. We documented each step meticulously, including the study selection and exclusion procedures. (presented in Fig. \ref{fig2}). 

\emph{Eligibility criteria:} This review will focus on the optimization and acceleration of LLMs, examining the most recent and widely utilized libraries, frameworks, and techniques in this field. To ensure focused analysis, strict eligibility criteria are applied. Only studies published between 2017 and December 2023 are considered, excluding publications not written in English and retracted papers. Additionally, studies are excluded if they are irrelevant to our SLR, or do not explicitly address ``Optimization'' ``Acceleration,'' or ``Large Language Models'' in their titles, abstracts, or keywords.

\emph{Information sources:} To ensure a comprehensive search for authentic studies, a variety of sources, including databases, websites, and tools, were employed. Digital libraries like IEEE Xplore, Web of Science, and Scopus alongside open access libraries like arXiv and dedicated tools like Zotero facilitated the data collection and reference management. The last search was conducted on May 25\textsuperscript{th}, 2024. Additionally, Rayyan and the researchrabbit.ai websites were utilized for data exploration and study selection.

\emph{Search strategy:} This systematic review leveraged two web-based AI tools, ResearchRabbit \cite{b32}, and Rayyan \cite{b33}, for both data collection and study selection. In all databases and websites, we were particularly interested in finding studies that focused on language modeling, particularly those that focused on LLM optimization and acceleration. We employed various queries in each source (see Table \ref{rosta.t1}) and exported the retrieved studies for import into Rayyan. Rayyan's AI capabilities facilitated both the selection of desired studies and the exclusion of irrelevant ones. 

\begin{table}
\caption{\textbf{Research queries executed}}
\label{rosta.t1}
\setlength{\tabcolsep}{3pt}
\begin{tabular}{|p{35pt}|p{180pt}|}
\hline
No. & Research Query\\
\hline
RQ 1 & LLM GPU Acceleration\\
\hline
RQ 2 & LLM GPU Optimization\\
\hline
RQ 3 & LLM Acceleration\\
\hline
RQ 4 & LLM Optimization \tablefootnote{The initial search query in arXiv with this RQ was broad, returning 440 studies, many irrelevant to our research. To refine the results and minimize the risk of bias, also ensure retrieval of high-quality, relevant papers, we employed the AND operator along with the title field within the search query specifically on arXiv.} \\
\hline
RQ 5 & Large Language Model GPU Acceleration\\
\hline
RQ 6 & Large Language Model GPU Optimization\\
\hline
\end{tabular}
\end{table}

\emph{Selection process:} The process of selecting which works to review in this study employed strict inclusion criteria. In this SLR we explore the techniques and methods that were primarily examined based on their focus on large-scale language modeling, including transformer-based models such as PLMs, LLMs, and even general NLP models. The Rayyan platform facilitated the selection process. Two stages were involved: initial screening using eligible and inclusion criteria, followed by author selection of the most relevant and impactful studies. Finally, the ``compute rating'' function in Rayyan was used, and the authors double-checked excluded studies for accuracy.

\emph{Data Extraction:} In this stage, we focused on extracting relevant data from selected studies. Our aim was to collect information on two key aspects:

\emph{Outcomes:} We were particularly interested in outcomes related to LLM optimization and acceleration. Specifically, we sought data on:
    \begin{itemize}
        \item Performance metrics: This could include metrics like perplexity \cite{b18}, BLEU score, ROUGE score \cite{hu2021lora}, or task-specific accuracy measures depending on the study's focus.
        \item Training time reduction: We looked for data on how different techniques impacted the time required to train LLMs.
        \item Resource usage: If studies reported resource (memory) usage changes with different optimization techniques, we collected that data.
    \end{itemize}
    
    We aimed to collect all relevant results within these outcome domains whenever possible. This included considering data from different measures, time points, and analyses reported by the study authors.

\emph{Additional variables:} In addition to the main outcomes, we also extracted data on the following aspects of the studies:

\begin{itemize}
    \item LLM architecture: The specific type of LLM architecture used in the study.
    \item Optimization techniques: Detailed description of the optimization techniques employed in the study.
    \item Hardware/Software platforms: The hardware and software platforms used for training, inference, serving, and evaluation.
\end{itemize}

\Figure[t!](topskip=0pt, botskip=0pt, midskip=0pt)[scale=0.5]{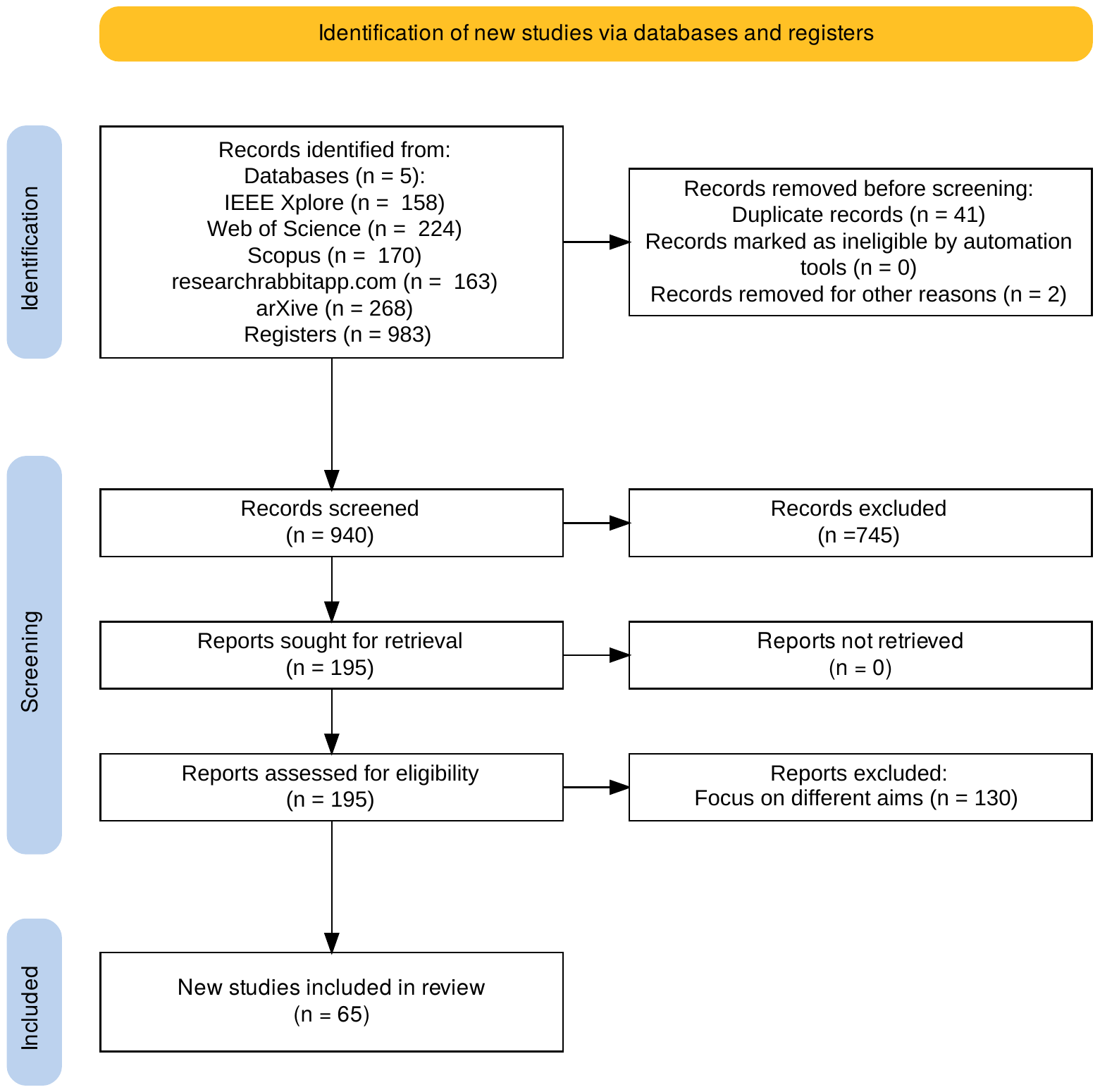}
{\textbf{The PRISMA 2020 flow diagram of the performed search.}\label{fig2}}

\emph{Data collection process:} ResearchRabbit is a web-based tool powered by AI that helps and guides researchers to find relevant studies in a variety of digital libraries and allows researchers to export retrieved results in a collection to reference managers tools (similar to Zotero). ResearchRabbit's search is powered by SemanticScholar and shows only the top 50 search results for a single query, aiming to maintain the research focus effectively \cite{b32}. Initially, we applied our queries to the ResearchRabbit website and then added the most relevant retrieved results to our collection. Following that, we applied the same queries in digital libraries like IEEE Xplore, Web of Science, Scopus, and arXiv (see Table \ref{rosta.t2}). The papers were reviewed on a case-by-case basis. Then, a precise summary of each paper was written. Finally, the interesting data that directly addressed the issues the papers attempted to address were extracted from the summaries.

\emph{Study Risk of Bias Assessment:}
In this SLR, we followed a meticulous process to assess the risk of bias in the included studies, adhering to best practices for ensuring the reliability and validity of our findings.

\emph{Automation Tools:}
\begin{itemize}
    \item We utilized Rayyan, an AI-powered tool, to facilitate the initial screening and selection process. Rayyan's AI capabilities helped in identifying potential biases and categorizing studies based on relevance and quality.
    \item ResearchRabbit was used for gathering relevant studies, which provided a focused list of top search results, aiding in maintaining the research scope effectively.
\end{itemize}

\emph{Reviewer Process: }   
    \begin{itemize}
    \item Each study was assessed by three independent reviewers. This approach helps to minimize subjective bias and ensures a more balanced evaluation.
    \item The reviewers independently examined each study based on predefined criteria including selection bias, performance bias, detection bias, attrition bias, and reporting bias.
\end{itemize}
\emph{Independent Review and Consensus:}
\begin{itemize}
    \item The reviewers worked independently during the initial assessment phase to ensure unbiased evaluations.
    \item After the independent assessments, the reviewers compared their findings. Any discrepancies or disagreements were resolved through discussion and consensus. 
\end{itemize}


We adhered to a rigorous and systematic approach to assess the risk of bias, which involved multiple independent reviewers and the use of validated tools. Automation tools such as Rayyan and ResearchRabbit played a crucial role in streamlining the screening and selection process, thereby enhancing the efficiency and accuracy of our assessments. By combining independent reviews, consensus discussions, and advanced AI tools, we ensured a robust and unbiased evaluation of the included studies.

\emph{Synthesis Methods:} To enable a comprehensive and insightful analysis of LLM optimization techniques across diverse contexts, a three-tiered categorization scheme will be employed. The initial categorization will consist of group studies based on the utilized LLM libraries/frameworks and the optimization techniques investigated.  Subgroups within these categories will be further established based on the specific type of LLM or the NLP task addressed by the studies. This method enables a highly detailed examination of how the effectiveness of optimization techniques varies across different LLM and NLP task configurations. Additionally, key findings from each individual study will be summarized in tables, including details like the optimization technique used, LLM type, NLP task addressed, achieved performance metrics, and the study's aims. Finally, a narrative synthesis will be conducted to analyze recurring themes across the studies. This thematic analysis will focus on the effectiveness of LLM libraries and optimization techniques in achieving performance improvements while considering resource constraints. It will also explore potential explanations for observed variations in effectiveness, with particular attention paid to factors like LLM size, resources used, and the NLP task addressed.

\emph{Reporting Bias Assessment and Certainty Assessment:} To minimize the risk of bias in our systematic review, we implemented a multifaceted strategy. First, to address reporting bias, we utilized Rayyan and ResearchRabbit, as AI-powered tools, during the initial screening and selection process. These tools can categorize studies based on relevance and quality and can help flag studies with characteristics suggestive of reporting bias, such as those focusing solely on positive outcomes. Second, to strengthen the certainty of our findings and minimize subjective bias, we implemented a multi-reviewer approach. Each study underwent independent assessment by three reviewers based on predefined criteria. This approach ensures a more balanced evaluation and reduces the influence of individual reviewer bias.

\begin{table}
\caption{\textbf{Studies retrieved per database / search engine}}
\label{rosta.t2}
\setlength{\tabcolsep}{3pt}
\begin{tabular}{|p{130pt}|p{85pt}|}
\hline
Database / Search Engine & Total\\
\hline
IEEE Xplore & 158\\
Web of Science & 224\\
Scopus & 170\\
ResearchRabbit & 163\\
arXiv & 268\\
\hline
\hline
Total & 983\\
\hline
\end{tabular}
\end{table}

\section{Language Modeling Development}
\label{sec:language_modeling_development}
Language modeling is a fundamental approach to enhancing the ability of machines to understand and process human language. It is a computational model that can learn and predict the possibilities of incoming (or missing) tokens \cite{b23}. The development of language models can be classified as follows (see Fig. \ref{fig3}):
\begin{itemize}
    \item \emph{N-gram language models,} like bigrams and trigrams, are basic methods that learn from the frequency of word sequences in text \cite{b34, b35}. However, their limited context window restricts their ability to capture long-range dependencies and understand the deeper semantic relationships between words.
     \item \emph{Markov assumption language models,} refers to those models that predict the next word based on the most recent in the context \cite{b23}. Both n-gram and Markov assumption language models are commonly used to improve task performance in NLP, and information retrieval (IR) \cite{b36}.
     \item \emph{Machine learning models,} these models investigate machine learning algorithms to enhance language comprehension. They are trained on extensive text corpora to discern patterns and relationships \cite{b37}. The adoption of machine learning in NLP introduced a more advanced methodology, enabling the creation of applications such as spam detection \cite{b38} and sentiment analysis \cite{b39}.
     \item \emph{Neural language models,} these models are developed based on NN for working with a sequence of data. They have a special ability of learning effective features for words or sentences. These studies \cite{b40, b41, b42} have initiated the use of language models for representation learning (beyond word sequence modeling), and show that these models have an important impact on the field of NLP \cite{b23, b43}.
     \item \emph{Transformer language models} refer to those models that leverage the capabilities of a deep learning architecture called Transformer to process and understand human language \cite{b44, b45}. These models achieved remarkable results by using \textit{``special attention mechanism''} to understand the relationship between words and sentences. These models capture context-aware representation instead of learning fixed word representations, first pre-training then fine-tuning according to specific downstream tasks \cite{b2, b8, b23, b46}. Transformer architecture has been used to build PLMs such as BERT \cite{b8}, GPT-2 \cite{b47}, and BART \cite{b48}. These models underwent training using bidirectional language models and specifically designed pre-training tasks applied to extensive unlabeled datasets. The growth in model size and data size has revolutionized the way we approach downstream tasks, enabling large-sized PLMs to achieve remarkable performance gains. These models exhibit unique characteristics compared to smaller PLMs, such as 330M-BERT and 1.5B-GPT-2, demonstrating exceptional abilities in solving complex tasks. As a result, LLM is the term used to refer to large-sized PLMs \cite{b46, b49, b50}.
\end{itemize}
\Figure[t!](topskip=0pt, botskip=0pt, midskip=0pt)[width=0.9\linewidth]{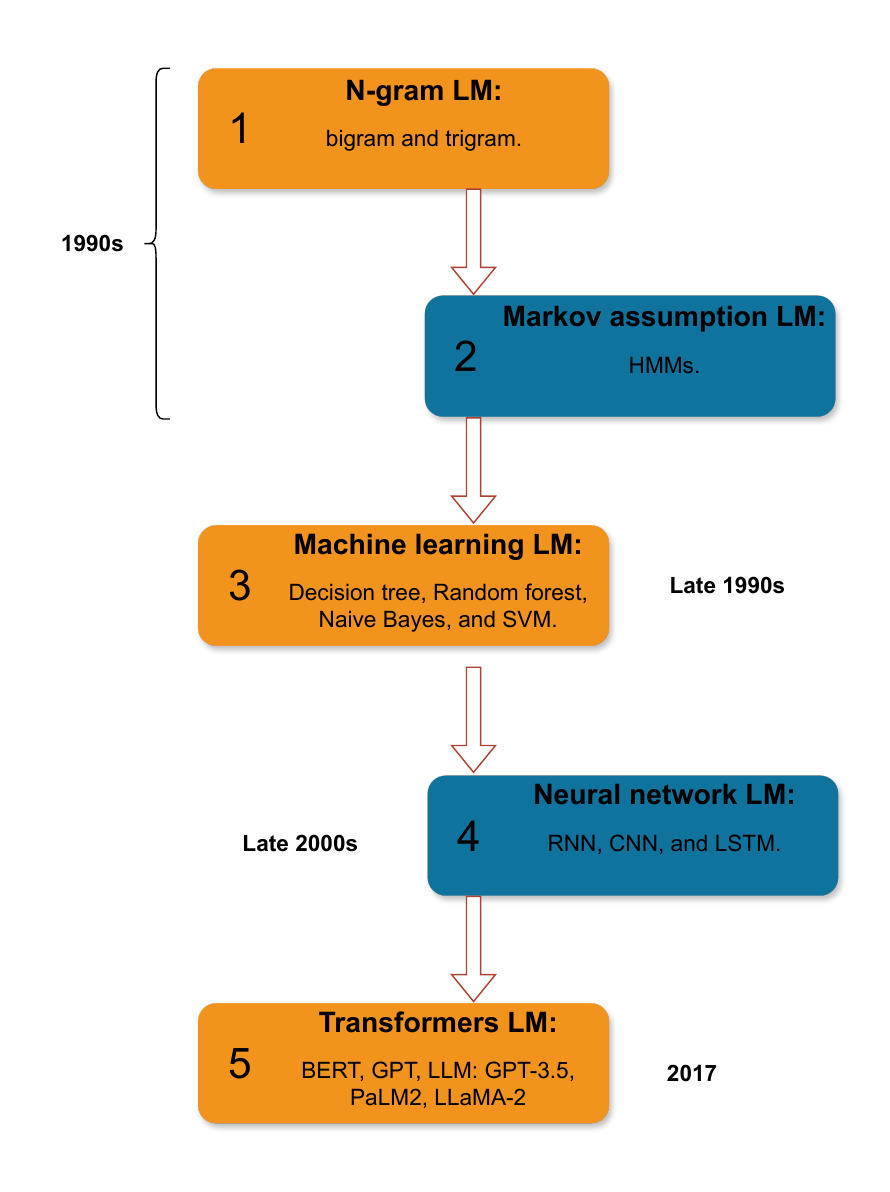}
{\textbf{Language model development}\label{fig3}}

\section{Machine Learning Models}
\label{sec:machine_learning_models}
The process of building, deploying, and managing a machine learning model involves three distinct phases: \emph{training, inference, and system serving}. Training is the foundation of machine learning, where a vast dataset of labeled data is used to develop a model that can identify patterns and relationships within the data. Inference is the application of the trained model, where new, unseen data is fed into the model to obtain predictions or classifications based on the learned patterns. System serving ensures the model's longevity and effectiveness in real-world applications, handling large volumes of requests, monitoring the model's performance, and providing continuous updates or modifications as needed \cite{b11, b18, b51}. In the section \ref{sec:frameworks_and_libraries}, we provide a categorization of the most recent frameworks and libraries utilized for LLMs optimization, structured into three primary classes: training, inference, and deployment and system serving (presented in Fig. \ref{fig4}). However, certain studies can be classified into two categories simultaneously, owing to their ability to handle multiple tasks, such as LightSeq2 (section \ref{subsubsec:lightseq2}), TurboTransformers (section \ref{subsubsec:turbotransformers}), and PetS (section \ref{subsubsec:pets}).

\Figure[t!](topskip=0pt, botskip=0pt, midskip=0pt)[width=0.5\textwidth]{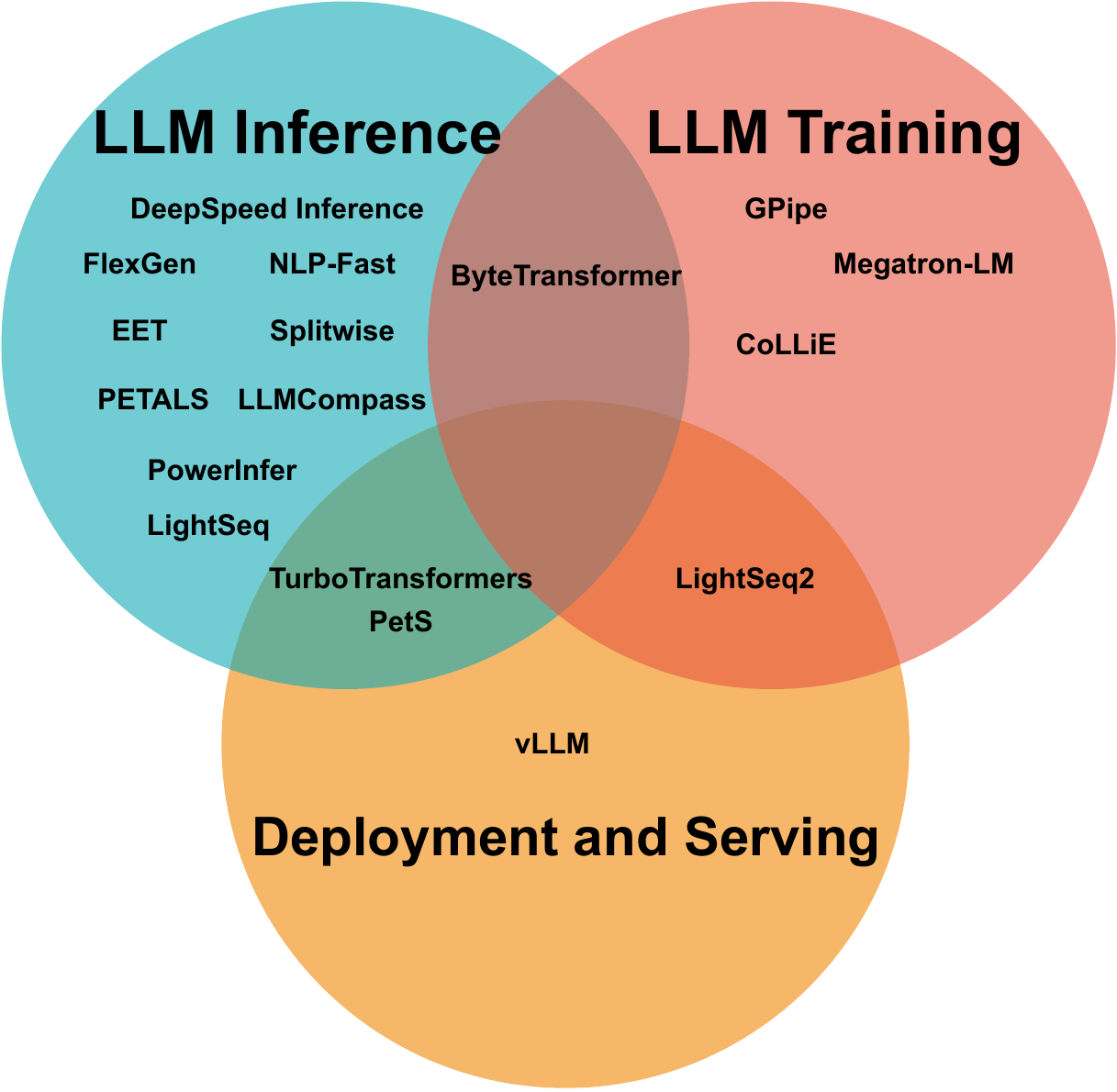}
{\textbf{LLM frameworks and libraries}\label{fig4}}

\begin{table*}[t]

\caption{\textbf{Summary of LLM training frameworks and libraries}}
\label{rosta.t3}

\begin{tabular}{p{60pt}p{170pt}p{240pt}}
\hline
Studies&
Aims&
Outcomes \\
\hline
GPipe \cite{b3} & A new \textbf{pipeline parallelism library} based on batch splitting. It is efficient, task independent, and works with different NN architectures.
& Investigated single 6-billion-parameter, 128-layer Transformer model on a dataset with 103 languages achieved better results than individually training 350-million-parameter.
\\
\hline
ByteTransformer \cite{b4}& A transformer \textbf{framework for GPU acceleration}, optimized for \textbf{variable-length inputs} in NLP problems. & Evaluated on an NVIDIA A100 GPU, for BERT-like transformers. Boosted fused MHA by 6.13$\times$ compared to PyTorch attention. Also, outperformed PyTorch, TensorFlow, Tencent TurboTransformer, Microsoft DeepSpeed, and NVIDIA FasterTransformer by 87\%, 131\%, 138\%, 74\%, and 55\%, respectively.
\\
\hline
Megatron-LM \cite{b18} & A deep learning \textbf{library for training LLMs with billions of parameters.} Offers a set of optimization methods for distributed training. & With 8.3 billion parameters trained on 512 NVIDIA V100 GPUs achieved 15.1 PetaFLOPs throughput. Also, achieving a perplexity of 10.8 on the WikiText103 benchmark and an accuracy of 66.5\% on the LAMBADA dataset.
\\
\hline
LightSeq2 \cite{b52} & Software \textbf{library that accelerates the training of transformer-based models within GPUs}. & Achieves significant speedups on a variety of NLP tasks. speedup of 308\% on the WMT14 English-German machine translation task compared to PyTorch.
\\
\hline
CoLLiE \cite{lv2023collie} & A \textbf{library for collaborative training of massive LMs}. It explores memory usage and throughput under different optimization methods, and investigates training techniques to improve the ability of a LM (LLaMA-65B) to follow user instructions. & Improves training efficiency for LLMs. Techniques like LoRA and AdaLomo specifically helped a large model (LLaMA-65B) follow instructions better, with an average score of 56.9, all without sacrificing overall performance.
\\
\hline

\multicolumn{3}{p{500pt}}{Bold text in the "Aims" column indicates the framework's primary area of specialization or the range of tasks it is designed to address. }\\

\end{tabular}
\end{table*}

\section{Frameworks and Libraries}
\label{sec:frameworks_and_libraries}
As most LLMs are designed based on Transformers, these models are a powerful type of neural network that have achieved SOTA results on a wide range of applications. To achieve these results the models are required to have a huge model size with hundreds of billions, even trillion of parameters. Training LLMs requires distributed training algorithms, which employ parallel processing techniques to efficiently train these massive models. To streamline distributed training, various optimization frameworks have been developed, providing tools and infrastructure for implementing and deploying parallel algorithms \cite{b23, b52, b53}. In this section, we will provide the most recent frameworks and libraries designed to overcome those limitations.

\subsection{LLM Training Frameworks and Libraries}
\label{subsec:llm_training_frameworks_and_libraries}
This section will delve into the objectives and outcomes of LLM frameworks and libraries employed in the training phase. Additionally, a summary of each framework/library will be provided individually (see Table \ref{rosta.t3}).

\subsubsection{GPipe}
\label{subsubsec:gpipe}
GPipe \cite{b3} introduces a novel pipeline parallelism framework based on batch partitioning. It divides each mini-batch example into smaller micro-batches, which are subsequently executed in sequence across the cells. During training, it employs synchronous mini-batch gradient descent, where gradients from all micro-batches are aggregated and applied to the model at the end of the mini-batch. GPipe has been shown to train two large-scale models: a convolutional model for image classification and a transformer model for machine translation. The convolutional model, AmoebaNet, was trained on 480$\times$480 input from the ImageNet 2012 dataset. To enhance its performance, the model width was expanded, and its parameters were scaled up to 557 million. The model achieved a top-1 validation accuracy of 84.4\%. Meanwhile, the transformer model a single 128-layer, 6-billion-parameter multilingual model trained across 103 languages, was also evaluated. GPipe achieved superior performance compared to training 350-million-parameter bilingual transformer big models individually across 100 language pairs. The model presents its efficiency by boosting the performance on a variety of devices, with the support of flexibility on any deep network architectures, utilizing the synchronous gradient decent, and ensuring consistent training regardless of the number of partitions.

\subsubsection{ByteTransformer}
\label{subsubsec:bytetransformers}
ByteTransformer \cite{b4} is a transformer framework for GPU acceleration with an efficient and high performance optimized for variable-length inputs in NLP problems. The framework uses an algorithm for overcoming the redundant computations on zero-padding tokens, and variable input length. Furthermore, the model proposed a fused Multi-Head Attention (MHA) to reduce the memory overhead of the intermediate matrix. This model manually optimizes the memory sizes of layer normalization by introducing bias and activation to maximize the overall system performance. It has been used by some famous applications including TikTok and Douying of ByteDance. The model was evaluated on an NVIDIA A100, focusing on the forward pass of BERT-like transformers, including BERT \cite{b8}, ALBERT \cite{b54}, DistilBERT, and DeBERTa. It showcased a significant improvement, enhancing the fused MHA mechanism by 6.13$\times$ compared to PyTorch attention. Additionally, ByteTransformer outperformed PyTorch, TensorFlow, Tencent TurboTransformer \cite{b11}, Microsoft DeepSpeed \cite{b5}, and NVIDIA FasterTransformer by 87\%, 131\%, 138\%, 74\%, and 55\%, respectively, in terms of the end-to-end performance of a standard BERT transformer.

\subsubsection{Megatron-LM}
\label{subsubsec:megatron_lm}
Megatron-LM \cite{b18} is a deep learning library for training LLMs efficiently and effectively.  It enables the library for the training of very large transformer models with billions of parameters. It offers a set of optimization methods for distributed training, it includes strategies like intra-layer model parallelism, and mixed-precision training. These optimization techniques significantly enhance training efficiency and speed, facilitating effective distributed training across multiple GPUs. Megatron-LM operates independently without requiring new compiler or library changes. This makes it orthogonal and complementary to pipeline model parallelism, allowing for seamless integration and flexibility within existing NLP frameworks.

The library has been shown to be highly effective for training LLMs. A Megatron-LM model with 8.3 billion parameters was trained on 512 NVIDIA V100 GPUs using 8-way model parallelism and achieved sustained performance of up to 15.1 PetaFLOPs across the entire application. This is significantly faster than previous approaches to training LLMs. Additionally, it has been shown to achieve SOTA results on several NLP benchmarks. A Megatron-LM model with 8.3 billion parameters achieved a perplexity of 10.8 on the WikiText103 benchmark. Also, it achieved an accuracy of 66.5\% on the LAMBADA dataset, which outperforms the previous SOTA of 63.2\%. 

\subsubsection{LightSeq2}
\label{subsubsec:lightseq2}
LightSeq2 \cite{b52} proposes a software library that accelerates the training of transformer-based models within GPUs. It is a system-level optimization while maintaining accuracy and training behavior. The system works with BERT (encoder), GPT (decoder), Transformer (encoder-decoder), and vision transformer. The system uses three techniques for improving training speed and efficiency. First (layer-specific kernels), after analyzing Transformer-specific layers in detail, rewriting the kernels with dependencies and other techniques to improve parallelism, and using small kernels to improve GPU utilization. Second (mixed-precision trainer), instead of applying batch updates to many individual full-precision updates, it applies batch updates to reduced-precision parameters. Finally, introduced an efficient memory management technique to minimize the need for frequent allocation and release calls. This strategy involves recycling the memory space of tensors that remain unused during the backward pass.

The system accelerates the entire training process for transformer models. LightSeq2 achieves significant performance improvement on a variety of NLP tasks, including machine translation, on the WMT14 English-German machine translation task, it achieved a 308\% speedup on the WMT14 English-German machine translation task compared to PyTorch.

\subsubsection{CoLLiE}
\label{subsubsec:CoLLiE}
CoLLiE \cite{lv2023collie} introduces a library designed to efficiently facilitate the collaborative training of LLMs using 3D parallelism \cite{b23} (Sections \ref{subsubsec:zero}, \ref{subsubsec:data_parallelism}, \ref{subsubsec:model_parallelism}), parameter-efficient fine-tuning (PEFT) methods, and optimizers.  The library demonstrated significantly improved training efficiency compared to existing solutions. The study empirically evaluates the correlation between model size and GPU memory consumption under different optimization methods and analyzes throughput. Additionally, the study investigates training methods to improve the abilities of the LLaMA-65B model, specifically focusing on following user instructions. Techniques like LoRA \cite{hu2021lora}, LOMO \cite{lv2023full} (Section \ref{subsubsec:fine_tuning}), AdaLomo \cite{lv2023adalomo}, and AdamW demonstrated success in boosting the model's instruction following capabilities without sacrificing its overall performance. Notably, LoRA and AdaLomo achieved impressive results, enabling the model to achieve an average score of 56.9.

\subsubsection{LLM Training Frameworks and Libraries: Challenges and Key Findings}
\label{subsubsec:training_challenges_and_keyfindings}

This section explores five prominent frameworks and libraries: GPipe \cite{b3}, ByteTransformer \cite{b4}, Megatron-LM \cite{b18}, LightSeq2 \cite{b52}, and CoLLiE \cite{lv2023collie}. Each offers unique functionalities to overcome limitations in LLM training.

\emph{Addressing Training Challenges:}
\begin{itemize}
    \item Distributed training: As LLMs grow complex, training them on a single device becomes impractical. Frameworks like Megatron-LM \cite{b18} and CoLLiE \cite{lv2023collie} employ distributed training algorithms that split the model across multiple GPUs, enabling parallel processing and faster training.
    \item Efficiency and speed: LightSeq2 \cite{b52} tackles training speed through system-level optimizations. It utilizes techniques like layer-specific kernels and mixed-precision training to enhance GPU utilization and reduce memory usage. Similarly, ByteTransformer \cite{b4} accelerates transformer models for variable-length inputs in NLP tasks.
    \item Memory management: Efficient memory allocation is crucial for LLM training. CoLLiE \cite{lv2023collie} overcomes memory constraints in LLM training by utilizing 3D parallelism to efficiently distribute memory across training machines and GPUs, enabling the training of large models even in resource limited environments.
    \item Fine-tuning and performance: CoLLiE \cite{lv2023collie} investigates methods to improve specific LLM capabilities, such as following user instructions. It explores parameter-efficient fine-tuning methods that enhance model performance in targeted areas without compromising overall functionality.
\end{itemize}

\emph{Key Findings:}
\begin{itemize}
    \item GPipe \cite{b3} demonstrates successful training of a large multilingual transformer model, achieving superior results compared to training individual smaller models.
    \item ByteTransformer \cite{b4} significantly outperforms existing frameworks in terms of performance for BERT-like transformers on various benchmarks.
    \item Megatron-LM \cite{b18} facilitates training of LLMs with billions of parameters, achieving SOTA on NLP tasks while offering high throughput.
    \item LightSeq2 \cite{b52} accelerates transformer model training by up to 308\%, showcasing substantial performance improvements.
    \item CoLLiE \cite{lv2023collie} introduces a library for collaborative LLM training, demonstrating improved efficiency and effectiveness in training large models like LLaMA-65B. It explores methods to enhance specific functionalities without impacting overall performance.
\end{itemize}

\begin{table*}[t]

\caption{\textbf{Summary of LLM inference frameworks and libraries}}
\label{rosta.t4}

\begin{tabular}{p{60pt}p{170pt}p{240pt}}
\hline
Studies&
Aims&
Outcomes \\
\hline
DeepSpeed \par Inference \cite{b5} & GPU-only solution, powerful and versatile system for \textbf{efficient transformer inference} at scale. & Boosts throughput by 1.5$\times$, minimizes latency by 7.3$\times$, empowers inference of 25$\times$-larger models at 84 TFLOPS.
\\
\hline
FlexGen \cite{b16} & An offloading engine model designed for high-throughput LLM inference by efficiently \textbf{utilizing limited resources} from GPU, CPU, and disk, employing various techniques to enhance efficiency. & \textbf{1)} Achieved 40$\times$ throughput speedup with 5000-second latency for batch size 64 or 2048 tokens. \textbf{2)} Achieved 79$\times$ throughput speedup with 12000-second latency for batch size 256 or 8192 tokens. \textbf{3)} Achieved 100$\times$ throughput speedup with 4000-second latency for batch size 144 or 4608 tokens using 4-bit quantization compression.
\\
\hline
NLP-Fast \cite{b55} & Accelerates the performance of large-scale \textbf{heterogeneous} NLP models.
& Evaluated a variety of NLP models and hardware platforms, including CPU, GPU, and FPGA. The throughput improved by up to 2.92$\times$, 1.59$\times$, and 4.47$\times$ over the baseline performance.
\\
\hline

TurboTransformers \cite{b11} & A lightweight, easy-to-use system that enables efficient deployment of transformer models for \textbf{online services}. & It introduces three innovative features:\textbf{ 1)} Efficient GPU-based batch reduction kernels for Softmax and LayerNorm. \textbf{2)} Sequence-length-aware memory allocation algorithm. \textbf{3)} New batch scheduler employing dynamic programming for optimal throughput on variable-length requests.
\\
\hline

PetS \cite{b56} & A \textbf{unified framework} for multitask PET serving in a single system. & Enables 26× more concurrent tasks and enhances serving throughput by 1.53$\times$ on Desktop GPUs and 1.63$\times$ on Server GPUs.
\\
\hline

PETALS \cite{b57} & A collaborative platform for \textbf{distributed inference} and fine-tuning of LLMs \textbf{over the internet}. & With an optimal hardware setup involving CPU RAM offloading via PCIe 4.0 and GPU pairs connected through PCIe switches, offloading 176B parameters takes 5.5 seconds in a regular setup and 11 seconds in a multi-GPU setup, with each GPU having 1 GB of memory per billion parameters and PCIe 4.0 throughput of 256 Gbit/s (or 128 Gbit/s behind a PCIe switch for two GPUs).
\\
\hline

LightSeq \cite{b58} & Inference library, addresses the need for efficient and convenient deployment of Transformer models in \textbf{online services}. & In machine translation benchmarks, consistently outperforms TensorFlow and FasterTransformer (FT), achieving up to 14$\times$ and 1.4$\times$ speedups, respectively.
\\
\hline

Easy and Efficient \par Transformer (EET) \cite{b59} & Library to \textbf{accelerate transformer inference}. & 
Compared against Fairseq, LightSeq, and Faster Transformer (FT) on 2080Ti and A100, EET achieves speedups of 4.48-20.27$\times$ and 4.30-27.43$\times$, respectively, over Fairseq. On 2080Ti, EET achieves a speedup of 0.82-2.46$\times$ over LightSeq for model sizes of 768 and 1024. Additionally, EET achieves speedups of 1.21-6.30$\times$ and 1.62-812$\times$ over FT v3.1 on 2080Ti and A100, respectively, and a speedup of 1.40-4.20$\times$ over FT v4.0 on A100. 
\\
\hline

Splitwise \cite{patel2023splitwise} & \textbf{Improve LLM inference efficiency by separating compute-intensive and memory-intensive phases} onto different machines with hardware specialization. & Achieve significant outcomes like up to 1.4$\times$ higher throughput at 20\% lower cost or 2.35$\times$ higher throughput with same cost and power consumption.
\\
\hline

Zhang \textit{et al}., \cite{zhang2023hardware} \par LLMCompass  & \textbf{Evaluate hardware design} for LLMs using LLMCompass library for recommending cost-effective hardware designs. & Achieve significant outcomes like accurate (average error 10.4\% for task time, 4.1\% for LLM tasks), simulates GPT-3 175B on 4$\times$ A100 GPUs in 16 minutes, identifies more affordable hardware designs.
\\
\hline

PowerInfer \cite{song2023powerinfer} & Build a \textbf{faster LLM inference engine} for consumer-grade GPUs. & Achieve significant speedups (up to 11.69$\times$ faster) by optimizing for hot neurons on GPU and cold neurons on CPU, while maintaining accuracy and approaching performance of high-end GPUs.
\\
\hline
\multicolumn{3}{p{500pt}}{Bold text in the "Aims" column indicates the framework's primary area of specialization or the range of tasks it is designed to address. }\\

\end{tabular}
\end{table*}

\subsection{LLM Inference Frameworks and Libraries}
\label{subsec:llm_inference_frameworks_and_libraries}
This section will introduce the LLM frameworks and libraries designed particularly for inference tasks, followed by a summary of each one (see Table \ref{rosta.t4}).

\subsubsection{DeepSpeed Inference}
\label{subsubsec:deepspeed_inference}
DeepSpeed Inference \cite{b5} presents a comprehensive system solution for efficient transformer inference. It has the potential to enable new and innovative applications of transformer models in cloud datacenters and other resource-constrained environments. The system consists of two main parts: DeepSpeed Transformer and ZeRO-Inference \cite{b1}. The model is a GPU-only solution that leverages a variety of optimizations to achieve SOTA (minimize) latency and (maximize) throughput for transformer models of all sizes. Specifically, in the first phase DeepSpeed Transformer uses tensor-slicing and inference-optimized pipeline parallelism to scale dense transformer models across GPUs. 

For sparse transformer models, it has developed a massive-GPU sparse transformer layer that can extend the scalability of Mixture-of-Experts (MoE) transformer layers to hundreds of GPUs. This is achieved through a combination of parallelism techniques and optimization strategies for communication. Then, DeepSpeed Transformer employs optimized sparse kernels to reduce the computational burden on a single GPU. ZeRO-Inference \cite{b1} is a heterogeneous solution that leverages GPU, CPU, and NVMe memory to enable massive transformer inference with limited GPU resources. 

It is particularly useful for inferring models that are too large to fit in GPU memory. It works by partitioning the model weights across multiple GPUs and offloading unused weights to CPU and NVMe memory. This allows ZeRO-Inference to infer models that are much larger than would be possible with GPU-only solutions. As a result, DeepSpeed Inference boosts throughput by more than 1.5$\times$ for throughput-oriented scenarios and minimizes the latency by more than 7.3$\times$ compared to existing solutions for latency orientation scenarios. It facilitates real-time inference at a trillion-parameter scale by utilizing hundreds of GPUs, marking an unparalleled achievement in terms of inference scale. This technology allows for the inference of models that are 25 times larger than what GPU-only solutions can handle, achieving a substantial throughput of 84 TFLOPS, which is over 50\% of the A6000 peak performance.  

\subsubsection{FlexGen}
\label{subsubsec:flexgen}
Accelerating LLM inference is achievable by using multiple high-end accelerator technologies, due to their high computational and memory requirements. FlexGen \cite{b16} study proposes an offloading engine model which focuses on using (resource-constrained devices) limited resources to reach high-throughput LLM inference. The engine is flexible for configuration using different hardware resources by aggregating memory and computation from the GPU, CPU, and disk. To optimize throughput within the search space, researchers developed a linear programming-based search algorithm, through it the model can find efficient patterns for saving and accessing tensors. It has a larger space of batch size options to choose from without sacrificing accuracy through using 4-bit to compress weights and attention cache without the need for retraining or calibration. The model’s efficiency has been experimented by using NVIDIA T4 (16 GB) GPUs for running OPT-175B. It significantly outperforms DeepSpeed Zero-Inference \cite{b1, b5} and Hugging Face Accelerate by enabling significantly larger batch sizes, often reaching orders of magnitude higher than its competitors. As a result, it can achieve significant speedups in throughput. On a single T4 GPU equipped with 208 GB CPU DRAM and a 1.5 TB SSD, input sequence length 512, and output sequence length 32: With a latency of 5,000 seconds, it (effective batch size 64) surpasses DeepSpeed Zero-Inference (batch size 1) by over 40$\times$, whereas Hugging Face Accelerate fails to complete a single batch. Furthermore, it can reach 69$\times$ higher throughput with a higher latency of 12000 seconds compared to baselines (effective batch size 256, or 8192 tokens in total). Finally, the model can achieve 100$\times$ higher maximum throughput (effective batch size 144, or 4608 tokens in total) with 4-bit quantization to compression and 4000 seconds by holding all weights in the CPU and getting rid of disk offloading. The model achieved these results by aggregating memory and computation from the GPU, CPU, and disk, and by using a number of techniques to improve efficiency, such as I/O schedule tasks, possible compression techniques, and distributed pipeline parallelism. FlexGen is a significant advancement in LLM inference, as it enables high-throughput generation on resource-constrained devices. This opens new possibilities for deploying and using LLMs in a wider range of applications.

\subsubsection{NLP-Fast}
\label{subsubsec:nlp_fast}
NLP-Fast \cite{b55} is a system that accelerates the performance of large-scale heterogeneous NLP models by identifying performance-critical operations and applying holistic model partitioning, cross-operation zero skipping, and model/config adaptive hardware reconfiguration. NLP-Perf, a performance analysis tool, collects performance data for NLP models and identifies performance-critical operations. Holistic model partitioning is a comprehensive optimization technique, which integrates three model partitioning approaches: partial-head update, column-based algorithm, and feed-forward splitting, to facilitate end-to-end model partitioning. Cross-operation zero skipping, skips zero or near-zero values across multiple operations, which can significantly reduce the amount of computation required, these two optimization can be executed on different hardware platforms. Model/config adaptive hardware reconfiguration, reconfigures the model architecture for the specific hardware platform that it is running on, which can further improve performance. NLP-Fast has been evaluated on a variety of NLP models and hardware platforms, including CPU, GPU, and FPGA. The evaluation results show that NLP-Fast can improve throughput by up to 2.92$\times$, 1.59$\times$, and 4.47$\times$ over the baseline performance on each platform.

\subsubsection{TurboTransformers}
\label{subsubsec:turbotransformers}
TurboTransformers \cite{b11} is a lightweight, easy-to-use system that enables efficient deployment of transformer models for online services. It achieves SOTA performance on GPU platforms by proposing three innovative features that distinguish it from other similar models: Firstly, proposes an efficient and parallel GPU-based batch reduction kernels algorithm for Softmax and LayerNorm. Secondly, proposes a sequence-length-aware algorithm for memory allocation to efficiently balance memory allocation and deallocation, this algorithm overcomes the problem of variability of the input sentences. Finally, applying the framework involves utilizing a novel batch scheduler that leverages dynamic programming to achieve the optimal throughput on variable-length requests. It is a lightweight and easy-to-use system that can be integrated into PyTorch code with just a few lines of code. This makes the model a very accessible option for researchers and practitioners who want to use transformer models for online services. 

\subsubsection{PetS}
\label{subsubsec:pets}
The existing large-scale transformer modes follow the pre-train-then-fine-tune paradigm, copying the entire model for each downstream task consumes a lot of storage. This approach is unsuited for multi-purpose serving. Parameter Efficient Transformers (PET) reduce the resource overhead. They share the pre-trained model among tasks and only fine-tune a specific portion based on the task parameters. Prior to PetS \cite{b56}, the serving systems did not have any mechanism to provide flexibility for PET task management, and also there is no available efficient method to serve queries to different task batches.  It is the first unified framework for multi-task PET serving in a single system. As a class of transformer models PETs have been designed to be more efficient in terms of both parameters and computation. Therefore, PETs are well-suited for deployment in resource-constrained environments. Conventional serving frameworks move data between the GPU and CPU memory when the GPU cannot hold all of the data for the tasks that are being processed. This reduces the throughput of the system. It has the potential to revolutionize the way that LLMs are served, making it possible to deploy and run LLMs on a wider range of devices and with lower resource requirements. This could make LLMs more accessible to a wider range of users and businesses. Pets framework is a flexible PET tasks management mechanism and a specialized PET Inference Engine (PIE) that allows both inter-task and inter-algorithm query-batching. It enables 26$\times$ more concurrent tasks and enhances serving throughput by 1.53$\times$ on Desktop GPUs and 1.63× on Server GPUs.

\subsubsection{PETALS}
\label{subsubsec:petals}
PETALS \cite{b57} emerges as a collaborative platform specifically designed for the distributed inference and fine-tuning of LLMs over the internet. It aims to overcome the limitations associated with existing approaches, offering a range of advantages. The platform focuses on achieving high performance by leveraging pipeline parallelism, effectively enhancing the efficiency of LLM inference and fine-tuning processes. Furthermore, it showcases scalability, demonstrating its capability to support a substantial number of users and accommodate large-scale LLMs. This adaptability is complemented by the provision of a flexible API, allowing users to tailor the inference and fine-tuning processes according to their specific requirements. The PETALS key feature is its emphasis on collaboration, providing a framework that enables multiple participants to actively engage in LLM inference and fine-tuning tasks collectively. The collaborative nature of PETALS contributes to its potential in democratizing access to LLMs, making them more accessible and valuable across a diverse range of applications. In summary, PETALS emerges as a promising platform with the potential to enhance the accessibility and utility of LLMs. It can offload a 176B parameter model in 5.5 seconds for a regular setup and 11 seconds for a multi-GPU setup. These results demonstrate PETALS's superior performance for running large models with limited resources.

\subsubsection{LightSeq}
\label{subsubsec:lightseq}
LightSeq \cite{b58} is a lightweight inference library, addresses the need for efficient and convenient deployment of Transformer models in online services. It utilizes a combination of GPU optimization techniques, including coarse-grained fused kernel functions, hierarchical auto-regressive search, and dynamic GPU memory reuse strategy, to achieve significant performance gains compared to TensorFlow and FasterTransformer (FT). It supports a wide range of models and search algorithms, encompassing BERT, GPT, Transformer, and Variational Autoencoders (VAEs), whereas seamlessly integrating with popular models like BERT \cite{b8}, RoBERTa \cite{b60}, GPT, VAEs, MT Transformer, and Speech Transformer. The library is user-friendly, with a serving system and CUDA implementations, enabling easy deployment of popular models online without code modification. It addresses the deployment challenges of resource-intensive sequence models, narrowing the performance gap between large models and the demands of online services. In machine translation benchmarks, it consistently outperforms TensorFlow and FasterTransformer (FT), achieving up to 14$\times$ and 1.4$\times$ speedups, respectively.

\subsubsection{EET}
\label{subsubsec:eet}
Easy and Efficient Transformer (EET) \cite{b59} offers a library designed to accelerate transformer inference. It encompasses a range of optimizations for transformer inference, spanning both algorithmic and implementation aspects. To address the inefficiencies of explicit matrix addition and masked attention, the study implements custom CUDA kernels. Also, to extend all kernels to support a larger model size up to 12288 and a longer sequence above 4096 the research proposes a new method called thread block folding. Furthermore, the study introduced a CUDA memory management mechanism aimed at minimizing memory usage for models of the size. EET evaluated against Fairseq, LightSeq, and Faster Transformer (FT), On both a 2080Ti and A100, EET achieves a speedup of 4.48-20.27$\times$ and 4.30-27.43$\times$, respectively, compared to Fairseq. On a 2080Ti, EET outperforms LightSeq \cite{b58} with a speedup of 0.82-2.46$\times$ for model sizes of 768 and 1024. EET attains a speedup of 1.21-6.30$\times$ and 1.62-812$\times$ over FT v3.1 on a 2080Ti and A100, respectively, and a speedup of 1.40-4.20$\times$ over FT v4.0 on an A100.

\subsubsection{Splitwise}
\label{subsubsec:Splitwise}
Splitwise \cite{patel2023splitwise} investigates inefficiencies in LLM inference, which relies on expensive GPUs. The analysis reveals two distinct phases in LLM inference: a compute-intensive prompt computation and a memory-intensive token generation phase. While existing methods optimize batching and scheduling, they underutilize compute resources during token generation. To address this, the study proposes separating these phases across different machines. This allows for hardware optimized for each phase: powerful machines for prompt computation and potentially older, more cost-effective machines for token generation. Splitwise facilitates communication between these machines using fast interconnects. This approach enables the design of clusters optimized for throughput, cost, or power consumption. The model achieves up to 1.4$\times$ higher throughput at 20\% lower cost or 2.35$\times$ higher throughput with the same cost and power consumption. This approach improves LLM inference efficiency by leveraging hardware specialization, leading to more cost-effective and power-efficient deployments.

\subsubsection{LLMCompass}
\label{subsubsec:LLMCompass}
Zhang \textit{et al} \cite{zhang2023hardware} propose LLMCompass, a library that efficiently evaluates hardware design for LLMs. LLMCompass considers various hardware options and identifies the optimal configuration for a specific task. The study also uses a cost model to recommend the most economical design. The library demonstrates high accuracy, with an average error of 10.4\% for predicting task execution time and 4.1\% for LLM tasks, compared to real hardware. Notably, the model can simulate running a massive LLM like GPT-3 175B on a powerful computer setup with 4$\times$ A100 GPUs in just 16 minutes. Leveraging LLMCompass, the study identified hardware designs that are more affordable than current options (e.g., using less powerful components or cheaper memory) while still offering good performance. These designs could make LLMs more accessible to a wider range of users.

\subsubsection{PowerInfer}
\label{subsubsec:PowerInfer}
PowerInfer \cite{song2023powerinfer} is a high performance inference engine designed to run LLMs efficiently on consumer-grade GPUs. It leverages the power-law distribution of neuron activation in LLMs, assigning frequently activated (hot) neurons to the GPU and input-specific (cold) neurons to the CPU. This hybrid approach significantly reduces the pressure on GPU memory and minimizes data transfers between CPU and GPU. Furthermore, PowerInfer incorporates adaptive predictors and neuron-aware sparse operators to optimize performance and maintain model accuracy. Evaluations demonstrate that PowerInfer on an NVIDIA RTX 4090 GPU achieves inference speeds up to 11.69$\times$ faster inference than systems like llama.cpp.It delivers an average token generation rate of 13.20 tokens per second, rivaling the performance of top-tier server-grade GPUs.

\subsubsection{LLM Inference Frameworks and Libraries: Challenges and Key Findings}
\label{subsubsec:llm_inference_challenges_and_keyfindings}
This section presents various frameworks and libraries designed to improve the efficiency of running LLMs. The following paragraphs discuss the challenges and key findings of the reviewed studies.

\emph{Challenges of LLM Inference:}
\begin{itemize}
    \item LLMs are computationally expensive due to their massive size and complex architecture.
    \item Traditional inference methods struggle to handle large models on resource-constrained devices.
    \item Balancing speed, accuracy, and resource utilization is crucial for deploying LLMs in real-world applications.
\end{itemize}

\emph{Key Findings:}
\begin{itemize}
    \item Hardware specialization: Splitwise \cite{patel2023splitwise} proposes separating compute-intensive and memory-intensive phases onto different machines with specialized hardware.
    \item Resource optimization: FlexGen \cite{b16} utilizes various techniques like I/O scheduling, compression, and distributed processing to efficiently use resources from CPU, GPU, and disk.
    \item Algorithmic optimizations: Libraries like EET \cite{b59} and LightSeq \cite{b58} implement custom algorithms and memory management techniques to accelerate inference on GPUs.
    \item Heterogeneous platforms: NLP-Fast \cite{b55} leverages different hardware platforms (CPU, GPU, FPGA) by identifying performance-critical operations and applying targeted optimizations.
    \item Distributed inference: PETALS \cite{b57} facilitates collaborative inference and fine-tuning of LLMs across a network, enabling scalability and efficient resource utilization.
    \item Efficiency gains: Several frameworks achieve significant performance improvements. DeepSpeed Inference \cite{b5} boasts throughput boosts of 1.5$\times$ and latency reductions of 7.3$\times$. FlexGen demonstrates even greater throughput gains, particularly on resource-constrained devices. Other frameworks like NLP-Fast \cite{b55}, TurboTransformers \cite{b11}, LightSeq \cite{b58}, and EET \cite{b59} show promising results in accelerating inference.

\end{itemize}

\subsection{LLM Deployment and Serving Libraries}
\label{subsec:llm_deployment_and_serving_libraries}
As mentioned in section \ref{sec:frameworks_and_libraries}, some of the frameworks and libraries are utilized for multiple purposes. Besides vLLM \cite{kwon2023efficient} (Section \ref{subsubsec:vllm}), the models used for deployment and serving purposes are mentioned in these sections LightSeq2 \ref{subsubsec:lightseq2}, TurboTransformer \ref{subsubsec:turbotransformers}, PetS \ref{subsubsec:pets}.
\subsubsection{vLLM}
\label{subsubsec:vllm}
vLLM \cite{kwon2023efficient} is a high performance system that efficiently handles LLMs at a large scale. The model tackles the memory limitations of existing LLM serving systems through a novel algorithm called PagedAttention (Section \ref{subsubsec:algorithmic_optimization}). PagedAttention splits the KV cache into manageable blocks, minimizing wasted memory and enabling efficient sharing across requests. vLLM is a distributed system that supports popular LLMs and even models exceeding single GPU memory. Evaluations present vLLM significantly improves throughput by 2-4$\times$ faster compared to existing systems, especially for complex tasks involving long sequences, large models, and intricate decoding algorithms. This makes vLLM a significant advancement for efficient LLM processing, enabling faster and more scalable LLM applications.

\subsubsection{LLM Deployment and Serving Libraries: Challenges and Key Findings}
\label{subsubsec:llm_deployment_challenges_and_keyfindings}
As explored in previous sections (Sections \ref{subsubsec:training_challenges_and_keyfindings} and \ref{subsubsec:llm_inference_challenges_and_keyfindings}) a variety of LLM frameworks exist that hold promise for deployment and serving applications. This section will discuss the key challenges and findings associated with LLM deployment and serving.

\emph{Challenges of LLM Deployment and Serving:}
\begin{itemize}
    \item Memory limitations: Large LLMs can easily overwhelm the memory capacity of a single GPU. This limits their deployment and serving for real-world applications.
    \item Scalability: Effectively handling multiple user requests simultaneously with large LLMs requires efficient scaling solutions.
    \item Variability of input: LLM performance can suffer when dealing with input sequences of varying lengths, requiring dynamic memory allocation strategies.
    \item Ease of deployment: Integrating complex LLM serving systems into existing workflows can be challenging for researchers and practitioners.

\end{itemize}

\emph{Key Findings:}
\begin{itemize}
    \item PagedAttention: This algorithm (introduced by vLLM \cite{kwon2023efficient}) breaks down the KV cache into manageable blocks, minimizing wasted memory and enabling efficient sharing across requests. This is a significant improvement for processing large LLMs.
    \item Efficient GPU utilization: TurboTransformers \cite{b11} utilize techniques like parallel GPU kernels and dynamic batch scheduling to optimize performance on GPUs. This translates to faster inference for transformer-based models.
    \item System-level optimizations: LightSeq2 \cite{b52} demonstrates how system-level optimizations within the training process can significantly improve training speed and efficiency for transformer models. This translates to faster deployment of LLMs in general.

\end{itemize}

These findings from vLLM \cite{kwon2023efficient}, TurboTransformers \cite{b11}, and LightSeq2 \cite{b52} offer promising solutions for overcoming challenges in LLM deployment and serving. By focusing on memory management, efficient GPU utilization, user-friendly tools, and co-optimization.

\Figure[t!](topskip=0pt, botskip=0pt, midskip=0pt)[scale=0.5]{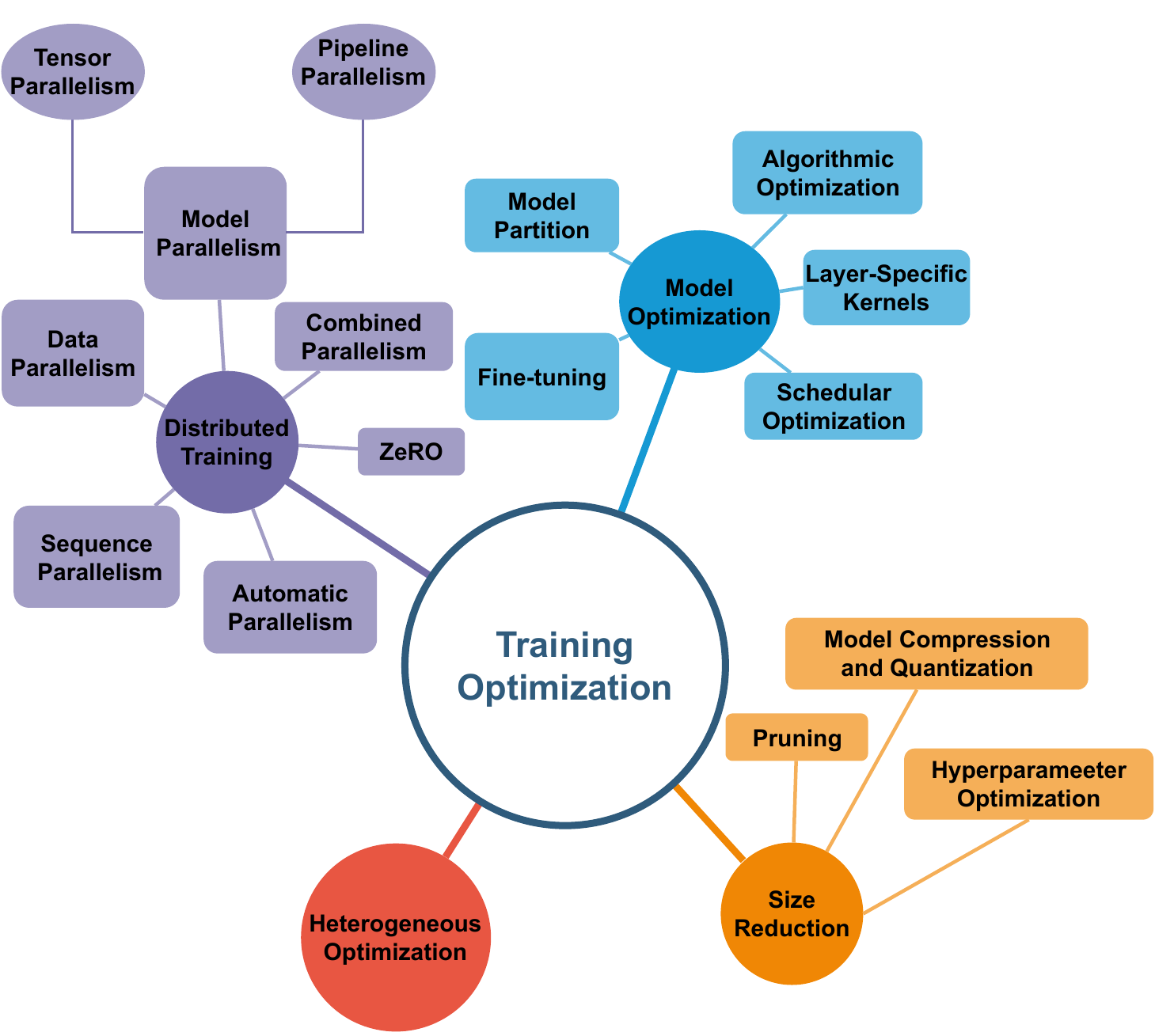}
{\textbf{Training optimization techniques}\label{fig5}}

\section{Training Optimization}
\label{sec:training_optimization}
Training optimization in LLMs involves improving the efficiency and effectiveness of the training process. This encompasses a range of techniques and strategies aimed at improving factors such as convergence speed, model generalization, and resource utilization. The goal of training optimization is to achieve the desired model performance with faster training times, reduced resource requirements, and improved overall training effectiveness. In this section, we will focus on model optimization, size reduction, distributed training, and heterogeneous training (Fig. \ref{fig5}).

\begin{table*}[t]

\caption{\textbf{Comparative analysis between different categories}}
\label{rosta.t5}
\scalebox{0.9} 
{
\begin{tabular}{p{60pt}p{120pt}p{120pt}p{90pt}p{110pt}}
\hline
Optimization&
Focus&
Performance&
Cost&
Scalability \\
\hline
Training Optimization & Focuses on accelerating the training process and minimizing resource usage within LLMs. This is accomplished through various techniques that enhance the efficiency of the training workflow. The aim is to achieve the same level of model performance in a reduced timeframe and with less computational power. & Techniques like SwapAdvisor \cite{b61} enables the training of models up to 12$\times$ larger than the standard GPU memory capacity while preserving substantial performance, ZeRO \cite{b17} achieves 10$\times$ speedup, train trillion parameter models (8$\times$ larger than existing models), Cramming \cite{b15} enables single-GPU LLM training in a one day, and Megatron-LM \cite{b68} outperforms ZeRO-3, achieving a remarkable 70\% improvement in performance for models with both 175 billion and 530 billion parameters. & Lower training costs due to faster training times, reduced memory usage, or enabling training on less powerful hardware. & Techniques like ZeRO \cite{b17} can scale to trillionn parameters, SparseGPT \cite{b67} processes very large models (OPT-175B, BLOOM-176B) efficiently, FlexFlow \cite{b72} improves parallelism efficiency by 2.5-10$\times$, ZeRO-Offload \cite{b19} enables training 10$\times$ larger models on the same hardware, and ZeRO-Infinity \cite{b1} Highly scalable for training models with trillions of
parameters. \\
\hline
Hardware Optimization & Systematically enhances the performance, efficiency, and functionality of computer hardware by addressing bottlenecks in hardware architecture, software, and operating systems. This approach can increase overall speed, reduce power consumption, and improve hardware reliability. Additionally, it enables more efficient use of hardware resources and allows for deployment on less powerful devices. & Techniques like FlexGen \cite{b16}, LightSeq2 \cite{b52}, TurboTransformers \cite{b11} improve performance (throughput, latency) for inference, potentially reducing operational costs. & Lower deployment costs by enabling inference on resource-constrained devices (CPUs, FPGAs) or requiring fewer servers for the same workload.& Techniques like TurboTransformers \cite{b11} and Splitwise \cite{patel2023splitwise} can potentially scale well on different hardware configurations.
\\
\hline

Scalability and Reliability & Improve the ability to train and run large models on distributed systems and handle potential hardware issues. & Techniques like PETALS \cite{b57} achieves faster inference for large language models, with an optimal setup inferring a 176 billion parameter model in 5.5 seconds. & Techniques like SWARM Parallelism \cite{b10} trains large models on unreliable, heterogeneous devices with low network bandwidth.& Techniques like ZeRO-Offload \cite{b19} enables large model training on single GPUs and scales to larger systems using model parallelism.
\\
\hline

\end{tabular}
}
\end{table*}

\subsection{Model Optimization}
\label{subsec:model_optimization}
Model optimization in LLMs refers to the process of improving the model's architecture, structure, or parameters to enhance its overall performance. We stated various techniques aimed at achieving better accuracy, efficiency, or both. Common model optimization strategies for LLMs include algorithmic optimization (section \ref{subsubsec:algorithmic_optimization}), layer-specific kernels (section \ref{subsubsec:layer_specific_kernels}), model partition (section \ref{subsubsec:model_partition}), fine-tuning (section \ref{subsubsec:fine_tuning}), and scheduler optimization (section \ref{subsubsec:scheduler_optimization}). 

\subsubsection{Algorithmic Optimization}
\label{subsubsec:algorithmic_optimization}
FlexGen \cite{b16} devised a linear programming-based search algorithm to optimize throughput within the search space. This model can identify efficient patterns for tensor saving and access.

Building on techniques for efficient model execution, SwapAdvisor \cite{b61}, proposes a novel approach to deep learning memory management, that enables the training and serving of large models despite limited GPU memory. Through smart swapping between CPU and GPU memory, it optimizes scheduling, memory allocation, and swap planning to maximize computational efficiency. This approach allows training models up to 12$\times$ beyond the usual GPU memory limit while maintaining significant performance. It stands as an innovative solution for deep learning with limited GPU resources.

NLP-Fast \cite{b55} employs algorithmic optimization techniques to enhance the performance of large-scale heterogeneous NLP models. One of the techniques is cross-operation zero skipping, which eliminates unnecessary computations by skipping zero or near-zero values across multiple operations. By leveraging these techniques, NLP-Fast can significantly improve the overall performance of NLP models on various hardware platforms.

ByteTransformer \cite{b4} was developed to address the challenges of redundant computations and memory overhead in transformer models,  it employs a combination of algorithmic optimizations and memory-efficient techniques, including a padding-free algorithm, fused MHA, and manually optimized memory sizes of layer normalization. These techniques effectively eliminate unnecessary computations, minimize memory footprint, and reduce the cost of accessing GPU global memory, leading to significant performance gains compared to other transformer frameworks.

Sheared LLAMA \cite{b62} model introduces a dynamic batch loading. This innovative algorithm efficiently adjusts the composition of sampled data within each training batch based on varying losses observed across different domains. The primary objective is to dynamically update the batch loading process to maximize learning efficiency, ensuring that the model achieves the reference loss approximately simultaneously across all domains. All training experiments have been done on a maximum of 16 Nvidia A100 GPUs (80 GB). 

GrowLength \cite{b63} enhances the pre-training of LLMs by gradually increasing the training length, it introduces an innovative method inspired by the principles of extending context windows during training. This innovative approach accelerates the pre-training phase of LLMs by dynamically and gradually extending the length of the training sentence. The primary benefit of this method lies in its adaptability and efficient resource utilization. It optimizes computational resources effectively, allowing models to process more tokens within a restricted time frame. Throughout the training process, the model incrementally increases the training length, resulting in reduced computational expenses and enhanced training efficiency. The experiments have been done in three different setups: 1) LLM128, in this setup the sentence length fixed of 128 tokens, totaling 0.36B. 2) LLM1024, sentence length was set to 1024 tokens, the same total tokens as LLM128, allowing direct runtime comparison. 3) GrowLength, in this experiment the method progressively grew from 128 to 1024 tokens, saving time with shorter lengths and enhancing performance at 1024 tokens. As a result, with equivalent tokens, LLM1024 required longer pre-training than LLM128. Using GrowLength led to a significant decrease in loss, and emphasized its computational efficiency and practical value in resource-constrained configurations.

PagedAttention \cite{kwon2023efficient} introduces another innovative approach to improve learning efficiency. This novel attention algorithm is inspired by virtual memory used in operating systems. The algorithm splits the KV cache into fixed-size blocks, similar to memory pages, reducing fragmentation and enabling efficient sharing across requests. This approach significantly improves memory utilization and allows for larger batch sizes.

\subsubsection{Layer-Specific Kernels}
\label{subsubsec:layer_specific_kernels}
LightSeq \cite{b58} (Section \ref{subsubsec:lightseq}) is a lightweight inference library instead of using a straightforward combination of the fine-grained GPU kernel functions in TensorFlow or PyTorch implementations, it utilizes a method known as coarse-grained fusion. This strategy mitigates the significant time costs associated with numerous kernel function launches and GPU memory I/O operations for intermediate results. Therefore, it achieves a significant reduction in the number of atomic kernel functions, leading to a remarkable performance boost compared to conventional TensorFlow approaches.

LightSeq2 \cite{b52} (Section \ref{subsubsec:lightseq2}) proposed a software library that accelerates the training of transformer-based models within GPUs. It is a system-level optimization while maintaining accuracy and training behavior. The library works with BERT (encoder), GPT (decoder), Transformer (encoder-decoder), and vision transformer. LightSeq2 uses three techniques for improving training speed and efficiency. The first technique used for increasing GPU utilization is layer-specific kernels technique. After analyzing Transformer-specific layers in detail, rewriting the kernels with dependencies and other techniques to improve parallelism, and using small kernels to improve GPU utilization.

\subsubsection{Model Partition}
\label{subsubsec:model_partition}
NLP-Fast \cite{b55} (Section \ref{subsubsec:nlp_fast}) accelerates the performance of large-scale heterogeneous NLP models by applying several techniques. It proposed holistic model partitioning as a solution for optimizing every operation in NLP models. This technique breaks down the model into smaller, more efficient submodels can be tailored for different hardware platforms. 

GPipe \cite{b3} (Section \ref{subsubsec:gpipe}) is an efficient, task-independent, and supports any deep neural network architecture that can be expressed as a sequence of layers. It can use different accelerators, each of which supports re-materialization. GPipe partitions the model across the accelerators, with each accelerator responsible for a sequence of layers (called a cell).

Megatron-LM \cite{b18} introduces a new method for training LLMs, which empowers the training of exceptionally large transformer models with billions of parameters within GPUs. Megatron-LM uses intra-layer model parallelism, a strategy that subdivides the model into smaller submodels capable of being trained separately.

\subsubsection{Fine-tuning}
\label{subsubsec:fine_tuning}
AlphaTuning \cite{b64} is a novel method specifically designed for large-scale pre-trained language models (PLMs). It combines the quantization of PLMs with fine-tuning, only a subset of quantized parameters is fine-tuned for the target task. This selective approach significantly decreases the overall memory footprint and the number of parameters to be trained. Despite these reductions, it maintains performance levels comparable to full fine-tuning across a diverse range of downstream tasks.

QFT \cite{li2023qft} proposes a novel framework designed for memory-efficient fine-tuning of LLMs. The model utilizes quantization techniques to significantly reduce memory usage during fine-tuning while preserving model performance. The framework adopts the Lion optimizer, known for its memory efficiency and compatibility with quantization, and the conversion of all model states into integers to minimize memory footprint. The study also features a specialized gradient flow and parameter update scheme tailored for quantized weights. Extensive evaluations show the framework's effectiveness, allowing fine-tuning of large LLaMA-7B models with less than 30 GB of memory on a single A6000 GPU a substantial reduction compared to standard methods while maintaining similar performance across various benchmarks.

LOMO \cite{lv2023full} is a novel technique for training LLMs on machines with limited GPU capacity. LOMO proposed a memory-efficient update method that greatly lowers memory consumption compared to traditional methods. This enables fine-tuning large models, such as those with 65 billion parameters, on consumer-grade GPUs like the RTX 3090. The study validates LOMO's efficiency through analyses of memory usage, performance testing, and benchmark task evaluations. Existing techniques like LOMO reduce memory usage but compromise performance.

AdaLomo \cite{lv2023adalomo} offers a better solution. It incorporates a key feature from the powerful AdamW optimizer (adaptive learning rate) but uses clever techniques to stay memory-friendly. This allows AdaLomo to match AdamW's performance on various tasks, making LLM training more accessible with less memory needed. On average, AdaLomo achieved scores of 30.8, 39.7, 51.0, and 56.9 on the LLaMA benchmark for models with 7B, 13B, 30B, and 65B parameters, respectively.

LoRA \cite{hu2021lora} is a method designed to adapt LLMs, such as GPT-3, for specific tasks, addressing the challenges of traditional fine-tuning. Instead of adjusting all pre-trained model weights, LoRA introduces trainable rank decomposition matrices into each layer of the Transformer architecture, significantly reducing the number of trainable parameters needed for downstream tasks. This approach reduces the number of trainable parameters by 10,000$\times$ and reduces GPU memory requirements by 3$\times$ compared to GPT-3 175B fine-tuned with Adam, while maintaining or improving model quality on benchmarks like RoBERTa, DeBERTa, GPT-2, and GPT-3. LoRA achieves higher training throughput with no added inference latency and facilitates efficient task-switching by sharing the pre-trained model and only optimizing the small low-rank matrices, thereby reducing storage and hardware costs. It is versatile and can be combined with other methods, applicable to any neural networks with dense layers. For GPT-3 175B, LoRA with 4.7M parameters achieves 73.4\% accuracy on WikiSQL, 91.7\% on MNLI-m, and Rouge-1/2/L scores of 53.8/29.8/45.9 on SAMSum, demonstrating its superior performance and efficiency.
 
\subsubsection{Scheduler Optimization}
\label{subsubsec:scheduler_optimization}
TurboTransformers \cite{b11} (Section \ref{subsubsec:turbotransformers}) introduces a novel sequence-length-aware batch scheduler that utilizes dynamic programming (DP) to optimize response throughput. This approach overcomes the limitations of traditional batch schedulers that struggle with varying input lengths. The model considers sequence length in batching decisions. The scheduler's core algorithm operates in \emph{O}(n\textsuperscript{2}) time complexity, making it efficient for real-time applications.

PetS \cite{b56} (Section \ref{subsubsec:pets}) introduces a unified framework aimed at enhancing multi-task PET serving efficiency. It comprises two main components: a flexible PET task management mechanism and a specialized PIE. Together, these components facilitate both inter-task and inter-algorithm query-batching, streamlining the processing of PET tasks. This approach optimizes resource utilization and enhances the efficiency of PET serving. The PET task scheduler efficiently schedules PET operations to run in parallel on the GPU, maximizing hardware utilization and performance. It dynamically assigns PET tasks to CUDA streams, considering both PET operator characteristics and system resource constraints. This lightweight online scheduling strategy effectively balances computational and memory-intensive tasks, leading to improved throughput and reduced latency in multi-task PET serving scenarios.

\subsection{Size Reduction Optimization}
\label{subsec:size_reduction_optimization}
Minimizing the size or complexity of LLMs is a crucial optimization technique known as size reduction optimization. This approach is essential for addressing challenges associated with memory demands, computational efficiency, and storage limitations. Size reduction optimization encompasses various techniques, including model compression and quantization (Section \ref{subsubsec:model_compression_and_quantization}), pruning (Section \ref{subsubsec:pruning}), and hyperparameter optimization (Section \ref{subsubsec:hyperparameter_optimization}).

Cramming \cite{b15} investigates the trade-offs involved in scaling down language model training, and investigates different parts of the training pipeline to identify the modifications that have the biggest impact on performance in a scaled-down setting. As a result, the research figured out that even under customized and constrained settings, the scaling laws \cite{b65} were almost true as it was observed for performance in large-compute settings. As a predictable outcome of these laws, it is a challenging task to perform downscaling. However, a smaller model architecture requires less computation power and allows to boost up the gradient computations, as a result the rates of the improved model within the time remain nearly unchanged. The study shared that doing modifications to the training methodology leverages scaling law to bring about enhancements by increasing the effective rate of gradient computations without sacrificing the model size. Two model setups have been analyzed: one utilizing a classical rtx2080ti GPU, and the other employing a modern rtxa4000 or rtxa6000 GPUs. Each setup was configured with 4 CPU cores and 32 GB of RAM. The paper proposes several modifications to the standard training pipeline to make it possible to train a language model on a single GPU in one day. As a result, each of these modifications has a direct impact on the model size reduction such as smaller model architecture, shorter training schedule, lower learning rate, mixed precision training, and specialized training library.

\subsubsection{Model Compression and Quantization}
\label{subsubsec:model_compression_and_quantization}
FlexGen \cite{b16} (Sections \ref{subsec:llm_inference_frameworks_and_libraries}, and \ref{subsubsec:algorithmic_optimization}) through a linear programming-based search algorithm identifies optimal patterns for tensor storage and retrieval. Furthermore, it employs 4-bit quantization to compress weights and attention cache without compromising accuracy, significantly reducing model size and memory footprint. These optimizations enable it to achieve impressive throughput gains compared to existing LLM inference systems. 

SWARM parallelism \cite{b10}, proposes a model for training a large model with unreliable heterogeneous devices with low network bandwidth by using dynamically generated, randomized pipelines instead of static pipelines dynamically instead of statically. The study incorporates 8-bit compression to minimize model size and facilitate training on resource-constrained devices with limited network bandwidth. This compression technique significantly reduces the amount of data that needs to be transferred between nodes during training, leading to improved efficiency and throughput.

QMoE \cite{frantar2023qmoe} is a compression and execution framework that reduces memory usage significantly. This is achieved through a scalable compression algorithm that shrinks trillion parameter MoEs down to less than 1 bit per parameter. This impressive compression is facilitated by a custom format specifically designed to work with bespoke GPU kernels, enabling efficient processing with minimal slowdowns. QMoE can compress the SwitchTransformer-c2048 model to under 160 GB (20$\times$ compression, 0.8 bits per parameter) with minimal impact on accuracy, achievable within a day on a single GPU. This enables the execution of trillion-parameter models on affordable commodity hardware, such as a single server with 4$\times$ NVIDIA A6000 or 8$\times$ NVIDIA 3090 GPUs, with less than 5\% runtime overhead compared to uncompressed inference. The framework reduces the model size from 3.2TB in bfloat16 to less than 160 GB, allowing efficient execution on commodity hardware and enhancing the practical adoption and research of MoE architectures.

AlphaTuning \cite{b64} is a compression-aware parameter-efficient adaptation method for large-scale PLMs. It combines the quantization of PLMs with fine-tuning, but only a subset of quantized parameters are fine-tuned for the target task. This significantly reduces the total memory footprint and the number of trainable parameters, while still achieving comparable performance to full fine-tuning on a variety of downstream tasks. It relies on binary-coding quantization, a technique that decomposes full-precision parameters into binary parameters alongside a distinct set of scaling factors. The model is evaluated across various PLMs and downstream tasks, and achieves comparable performance to full fine-tuning, even at low bitwidths. While it was applied to GPT-2 and OPT, it achieved a compression ratio of over 10 times under 4-bit quantization and a reduction in the number of trainable parameters by over 1,000-fold, while still achieving competitive performance on a variety of downstream tasks. 

GPTQ \cite{b66} proposes a new highly accurate and highly efficient post-training quantization method based on approximate second-order information which is called a new one-shot weight quantization. This model reaches a level that is considered acceptable to precisely quantize models to 3 or 4 bits per parameter, it requires a few hours at most to run on a model that has hundreds of billions of parameters. The model experimented on both OPT-175B and BLOOM-176B it took approximately 4 GPU hours by reducing the bitwidth down to 3 or 4 bits per weight, with minimal loss of accuracy compared to the uncompressed baseline. Compared to previous one-shot quantization methods the model achieves more than twice the compression without sacrificing accuracy. Also, within the method for first-time models with 175 billion parameters can execute inside a single GPU for generative inference. The results show that these enhancements can boost performance by up to 3.25$\times$ while using high-end GPUs (NVIDIA A100) over FP16 and reach 4.5$\times$ and up to 4.5$\times$ while using more cost-effective GPUs (NVIDIA A6000). This model can also achieve robust accuracy results even using an extreme quantization regime, while the weights are quantized to 2-bit or ternary quantization level.

FPTQ \cite{li2023fptq} also proposes a novel post-training quantization technique to address the deploying LLM challenge. This technique effectively compresses LLMs into a format using 4-bit weights and 8-bit activations (W4A8). This approach achieves SOTA performance on popular LLMs like BLOOM\cite{b69}, LLaMA\cite{b13}, and LLaMA-2 without requiring further fine-tuning. FPTQ offers a significant advantage by optimizing both memory usage and computation efficiency during the inference stage without sacrificing accuracy. This technique simplifies the deployment process for LLMs and makes them more practical for real-world use. The model was validated on various datasets, including LAMBADA, MMLU, and a set of Common Sense QA tasks. The researchers compared the model's performance to an existing technique called LLM-QAT (LLM-Quantization-Aware Training). However, limited data availability for LLM-QAT restricted the comparison to the Common Sense QA dataset. On this task, FPTQ achieved results closer to the FP16 compared to LLM-QAT. While the analysis was only possible for 7B and 13B parameter LLaMA models due to data limitations, FPTQ consistently performed better across all subsets of the dataset. This is evidenced by the average scores: 73.38 and 76.81 for LLaMA-7B and 13B, respectively. These findings suggest that FPTQ is an effective approach for LLM quantization.

Norm Tweaking \cite{b53} method introduces a novel technique in quantization, specifically for LLMs. While existing quantization methods like GPTQ \cite{b66} achieve acceptable 4-bit weight-only quantization, attempts at lower bit quantization often lead to significant performance degradation. It introduces a strategy to rectify the quantized activation distribution, restoring accuracy for LLMs. The method involves generating calibration data and applying channel-wise distance constraints to normalization layer weights. Experiments show significant improvements in both weight-only quantization and joint quantization of weights and activations, achieving high accuracy even at 2-bit quantization. It offers a practical solution for reducing computational and storage costs in LLMs while maintaining performance. 

FineQuant \cite{kim2023finequant} introduces an innovative weight-only quantization technique that significantly decreases memory usage and speeds up LLM inference with minimal quality loss. Key features of this technique include utilizing pre-trained model weights without further fine-tuning, applying adaptive granularity quantization to minimize accuracy loss, and implementing an efficient GPU processing approach. Tested on large-scale models like OPT-175B, FineQunat demonstrates minimal accuracy loss, achieves up to 3.65$\times$ higher throughput with the same number of GPUs, and reduces resource demands.

PETALS \cite{b57} (Section \ref{subsubsec:petals}) is a collaborative platform for distributed inference and fine-tuning of LLMs over the internet. To enhance efficiency, quantization techniques have been employed to store a higher number of parameters per GPU, thereby decreasing the need for consecutive devices and communication rounds and use 8-bit precision to compress the weights, reducing the nodes required to store all layers. To achieve more efficient data transfer between pipeline stages, dynamic blockwise quantization is utilized. It utilize 8-bit mixed matrix decomposition for matrix multiplication allows the model to quantize the weights to 8-bit precision, significantly reducing the memory footprint compared to 16-bit weights.

QFT \cite{li2023qft} (Section \ref{subsubsec:fine_tuning}) addresses memory limitations during fine-tuning LLMs by introducing a novel quantization framework. It converts all model states into integers to minimize memory footprint and employs the Lion optimizer for its memory efficiency and compatibility with quantization. Additionally, the framework incorporates a specialized scheme for handling quantized weights during training. 

QuantEase \cite{behdin2023quantease} is a framework for post-training quantization of LLMs that enhanced their deployment efficiency. The framework addresses the challenge of layer-wise quantization by optimizing each layer individually, utilizing Coordinate Descent (CD) to achieve high quality solutions efficiently without complex matrix operations. The framework includes an outlier-aware variant that maintains crucial ``outlier'' weights in full precision to enhance accuracy. Demonstrating SOTA performance, QuantEase significantly improves perplexity and zero-shot accuracy compared to existing methods like GPTQ\cite{b66}, with up to 15\% relative improvement. Efficient linear algebra optimizations allow for the quantization of large models such as Falcon-180B on a single GPU in under 3 hours. The outlier-aware variant supports near or sub-3-bit quantization with minimal accuracy loss, outperforming methods like SpQR by up to two times in perplexity reduction.

LLM-Pruner \cite{ma2023llm} (Section \ref{subsubsec:pruning}) compresses LLMs by removing non-essential parts based on gradient information while keeping their functionality. This significantly reduces model size with minimal accuracy loss, achieved through fine-tuning with a small amount of data.

\subsubsection{Pruning}
\label{subsubsec:pruning}
SparseGPT \cite{b67} framework has developed an efficient and precise post-training pruning technique for significantly reducing the size of large-scale GPT-family models. This method achieves at least 50\% sparsity in a single step, without requiring retraining. Remarkably, it enables the processing of the largest open-source models, such as OPT-175B and BLOOM-176B, in less than 4.5 hours. It makes the model achieve 50-60\% unstructured sparsity with a negligible increase in perplexity and removes more than 100 billion weights with minimal impact on accuracy. The study demonstrates that the parameterization of massive GPT models enables pruning without relying on the gradient information. It highlights that sparse models with comparable accuracy to dense models can be identified within the ``close neighborhood'' of the dense models. The study's findings reveal that sparse models achieve performance very similar to the dense models. The study also shows that it is easier to prune larger models: for a fixed sparsity level, the accuracy drop for larger sparse models is smaller, to the point where there is practically no accuracy decrease when reaching 50\% sparsity, to the point where reach 50\% sparsity does not result in any noticeable accuracy decrease on the largest models. 

Sheared LLAMA \cite{b62} is used to reduce the size of the LLaMA2-7B model to 1.3B and 2.7B parameters, and it performed better than other open-source models of the same size on a variety of downstream and instruction tuning evaluations. LLM-shearing also requires only 3\% of the computing resources to train as the same models trained from scratch. One of the main steps of Sheared LLAMA is a novel pruning algorithm that can prune a source model to any specified target architecture. The algorithm is an extended version of CoFiPruning that allows the source model to be pruned to any specified target architecture, based on the desired model size and performance requirements. Pre-trained models are typically well-optimized to balance expressivity and inference efficiency, therefore these configurations are used as the target architectures.

LLM-Pruner \cite{ma2023llm} introduces a framework for compressing LLMs in a task-agnostic way while minimizing the need for the original training corpus. The framework uses structural pruning to remove non-critical parts of the model based on gradient information. The pruned models' performance is recovered using LoRA \cite{hu2021lora} tuning, which takes just 3 hours and 50K data samples. Experiments on LLaMA \cite{b13}, Vicuna \cite{peng2023instruction}, and ChatGLM \cite{zeng2022glm} show that the compressed models maintain 94.97\% of their original performance even after removing 20\% of parameters. However, higher pruning rates lead to significant performance drops and incoherent sentence generation.

\subsubsection{Hyperparameter Optimization}
\label{subsubsec:hyperparameter_optimization}
Selecting the right hyperparameters is essential for developing effective LLMs, as these parameters significantly influence the model's convergence speed, generalization ability, and overall performance in various language tasks. Whereas often an iterative and computationally demanding process, hyperparameter optimization is crucial for achieving optimal model performance. Cramming \cite{b15} employs a lower learning rate to stabilize the training process and prevent overfitting, enabling effective model training within limited computational resources. 

\subsection{Distributed Training}
\label{subsec:distributed_training}
Distributed training refers to the process of training LLMs across multiple computing devices or processing units. This approach harnesses the power of parallelism to distribute the computational burden, enabling faster training of large models with millions or even billions of parameters. Distributed training is crucial for managing the massive datasets and computational demands associated with cutting-edge LLMs. 

\subsubsection{Data Parallelism}
\label{subsubsec:data_parallelism}
Data parallelism is a parallel training technique that replicates the entire model across multiple GPUs or devices and distributes the training data among them. Each device handles a portion of the data, performs forward and backward propagation, and computes gradients independently. These gradients are then aggregated across all devices to update the global model parameters. It is a fundamental and widely used technique for improving the training throughput of deep learning models. Its simplicity, scalability, and effectiveness make it a valuable tool for researchers and practitioners in the field of machine learning \cite{b14, b23, b68, b69}.

\subsubsection{Model Parallelism}
\label{subsubsec:model_parallelism}
The model parallelism can be classified into two groups tensor parallelism (section \ref{paragraph:tensor_parallelism}) and pipeline parallelism (section \ref{paragraph:pipeline_parallelism}).

\paragraph{Tensor Parallelism}
\label{paragraph:tensor_parallelism}
Tensor parallelism involves partitioning a tensor across an array of devices, necessitating a distributed matrix-matrix multiplication algorithm for mathematical computations. Using the tensor parallelism reduces the response time for individual queries \cite{b14, b16}. Megatron-LM introduced 1D tensor parallelism (Section \ref{subsubsec:megatron_lm}) which partitions the linear layer within the Transformer architecture along either the row or column dimensions. Within Megatron-LM, tensors are broken down into a single dimension \cite{b14, b18}.

\paragraph{Pipeline Parallelism}
\label{paragraph:pipeline_parallelism}
FlexGen \cite{b16} (Section \ref{subsubsec:flexgen}) utilizes pipeline parallelism to distribute an \emph{l}-layer LLM evenly across \emph{m} GPUs, enabling parallel execution of all layers. Each GPU executes the same sequence of operations, essentially reducing the problem to training an \emph{n/m-layer} transformer on a single GPU. This approach leverages the existing policy search algorithm developed for single-GPU training. In order to implement micro-batch pipelining, a new repetition statement (for-loop) is used within the applied algorithm effectively merging the iteration-level pipeline parallel execution schedule with a single-device offloading runtime.

PETALS \cite{b57} (Section \ref{subsubsec:petals}) utilizes pipeline parallelism to efficiently distribute the computation of LLMs among multiple servers. Servers are organized into a chain, with each server responsible for executing a portion of the model pipeline. This approach enables efficient parallel processing, improving the overall performance of inference and fine-tuning tasks. 

GPipe \cite{b3} (Section \ref{subsubsec:gpipe}) employs a novel pipeline parallelism algorithm based on batch splitting, where mini-batch examples are divided into smaller micro-batches and sequentially executed across cells during training. The model utilizes synchronous mini-batch gradient descent, accumulating gradients from all micro-batches and applying them to the model at the end of the mini-batch. The efficiency of the model is demonstrated through the successful training of large-scale models, including a convolutional model (AmoebaNet) for image classification and a transformer model for machine translation. The model showcases its flexibility across various deep network architectures, achieving superior results and consistent training performance on diverse devices.

DFX \cite{b70} is a low-latency multi-FPGA appliance for accelerating transformer-based text generation. It uses model parallelism to split the transformer model across multiple FPGAs. This allows each FPGA to process a different part of the model in parallel, thereby accelerating the overall text generation process. Also, it uses an efficient network to interconnect the FPGAs and reduce communication overhead. The network uses a ring topology to minimize communication overhead. This model utilized four Xilinx Alveo U280 FPGAs and evaluated its performance on the GPT-2 language model. It demonstrated a 5.58$\times$ acceleration in speed and a 3.99$\times$ enhancement in energy efficiency compared to four NVIDIA V100 GPUs. In addition to its performance and energy efficiency benefits, this solution proves to be more cost-effective than GPU-based alternatives. Moreover, it offers an 8.21$\times$ cost advantage over a GPU appliance delivering similar performance levels.

\subsubsection{Combined Parallelism}
\label{subsubsec:combined_parallelism}
Narayanan \textit{et al}., in \cite{b68} proposed a new technique called PTD-P for training LLMs on GPU clusters. PTD-P combines pipeline parallelism, tensor parallelism, and data parallelism to achieve high computational performance and graceful scaling. Data parallelism divides the training data into smaller batches, which are then processed in parallel on all the GPU servers. This allows PTD-P to achieve faster training by leveraging the parallel computing capabilities of the GPU cluster. Also, GPipe \cite{b3}, and ZeRO \cite{b17} (section \ref{subsubsec:gpipe}, and \ref{subsubsec:zero} respectively) are other examples of combined parallelism.

\subsubsection{ZeRo}
\label{subsubsec:zero}
ZeRO \cite{b17} proposed solutions to overcome the limitations of existing methods and efficiently train large models. While using existing systems the memory consumption can be classified into two main parts which are model states, and residual states. Most of the memory capacity is used by model states (such as momentum, variance in Adam, gradients, and parameters) while working with large models. The rest part of the memory is occupied by residual states (such as activation, temporary buffers, and unusable fragmented memory). For applying optimization in both model state memory and residual state memory, efficiently training models of such colossal sizes is crucial as they grow from billions to trillions of parameters. The study introduces a novel memory optimization technique aimed at substantially improving training speed, and with approach enables scaling the model size in proportion to the number of devices while maintaining high efficiency. Leveraging the latest hardware, this model can scale to over 1 trillion parameters by carefully evaluating communication volume and memory capacity requirements, this boosts memory efficiency for model states. For optimizing model state memory, which occupies most of the memory during training, the study introduces ZeRO-DP, ZeRO-powered data parallelism which has three main optimization stages: in the first stage, only the optimizer states are partitioned; in the second stage, both optimizer states and gradients are partitioned; and in the final stage, all three model states are partitioned. This results in a significant boost in memory efficiency. The rest of the memory consumed by residual states could become a secondary memory bottleneck. The study overcame this problem by three factors: Firstly using activation partition to optimize activation memory by locating and deleting activation replication in existing MP (model parallelism), and when appropriate offloads activations to CPU. Secondly, keeping the balance of memory and computation efficiency by introducing appropriate size temporary buffers to strike. Finally, during the training process, memory becomes fragmented because tensors have varying lifetimes. The lack of contiguous memory, resulting from this fragmentation, can lead to memory allocation failures, even when there is sufficient free memory space available. To address this problem, ZeRO-R takes a proactive approach by effectively handling memory based on the distinct lifetimes of tensors, thereby preventing memory fragmentation. Remarkably, this model achieves a throughput of 15 Petaflops when training models with over 100 billion parameters, demonstrating super-linear speedup on 400 GPUs. It indicates an 8$\times$ increase in model size and a 10$\times$ increase in performance compared to recent SOTA models.

\subsubsection{Sequence Parallelism}
\label{subsubsec:sequence_parallelism}
Sequence parallelism \cite{b14, b71}, is a novel approach proposed to efficiently train Transformers with longer sequences on GPUs. It addresses the quadratic memory requirements of self-attention in Transformer models. Unlike traditional methods, it does not require a single device to handle the entire sequence. By splitting sequences into chunks and distributing them across devices, it achieves effective training with infinitely long sequences. It introduces Ring Self-Attention to enhance the process, demonstrating superior performance in batch size and sequence length compared to tensor parallelism, handling sequences over 27$\times$ longer than existing methods.

\subsubsection{Automatic Parallelism}
\label{subsubsec:automatic_parallelization}
The automatic selection and parallelization strategies as the latest advances in parallel training demonstrated by FlexFlow \cite{b72} and Alpa \cite{b23, b73}. Alpa is an automated system that generates execution plans for distributed model-parallel training. It's an architecture that can automatically derive efficient parallel execution plans at each parallelism level. It is different from specialized systems as it can handle models with heterogeneous architectures and models without manually designed plans. However, it is not hardware-aware and does not consider network topology. Also, it does not search for activation checkpointing, which could lead to suboptimal results. Alpa has been evaluated on large models training with billions of parameters. The model's performance has been compared with the SOTA systems such as Megatron-LM \cite{b18} and DeepSpeed \cite{b5}, on an Amazon EC2 cluster with 64 GPUs. It presents the similar training performance as Megatron-LM on GPT models and outperforms DeepSpeed on GShared MoE models with up to 9.7$\times$ speedup. Moreover, it generalized well to models without manual strategies and demonstrated 80\% liner scaling efficiency on Wide-ResNet. The results presented that Alpa’s performance in training large models efficiently and its ability to generalize to diverse models.

\subsection{Heterogeneous Training}
\label{subsec:heterogeneous_training}
ZeRO-Offload \cite{b19}, aims to democratize large-scale model training, making it accessible to a wider audience. It achieves this by using a single GPU to train models with over 13 billion parameters, eliminating the need for data scientists to modify the models or sacrifice computational efficiency. The study introduces ZeRO-Offload, a novel heterogeneous deep learning (DL) training technology. The model leverages both CPU memory and compute for offloading and offers an efficient scaling path on multiple GPUs through collaboration with ZeRO-powered data parallelism \cite{b17}. Through first-principle analysis, the study asserts that the model provides an optimal solution, maximizing memory savings while minimizing communication and CPU compute overhead for large model training.

ZeRO-Infinity \cite{b1} introduces an innovative system technology that enables the model scaling on constrained resources. It achieves this without the need for extensive model code modifications by harnessing the power of GPU, CPU, and NVMe memory.  The model made up of five innovative technologies: 1) \textit{infinity offload engine}, this technique uses simultaneous exploitation of GPU, CPU, and NVMe memory, as well as GPU and CPU compute to fully leverage heterogeneous architecture on modern clusters, 2) \textit{memory-centric tiling}, handle extensive operators without necessity of model parallelism, 3) \textit{bandwidth-centric partitioning}, is employed to make the most of the aggregate memory bandwidth across all parallel devices, 4) \textit{overlap-centric design}, is implemented to enable the simultaneous execution of compute and communication tasks, 5)\textit{ ease-inspired implementation}, to prevent the need for extensive model code refactoring. SWARM Parallelism \cite{b10} (section \ref{subsubsec:model_compression_and_quantization}) introduced a model aimed at training large models efficiently, particularly on unreliable heterogeneous devices with limited network bandwidth. Instead of employing static pipelines, the model utilizes dynamically generated and randomized pipelines to adapt to varying conditions. This allows each device to share its results with any other device that is responsible for the next stage of the pipeline. This enables devices with high performance to process inputs from multiple predecessors, distribute their results across multiple weaker peers, and rebalance the workload in case of failure to improve utilization. 

NLP-Fast \cite{b55} (Section \ref{subsubsec:nlp_fast}) is a system designed to enhance the performance of large-scale heterogeneous NLP models by pinpointing the most resource-intensive operations and employing a combination of techniques: holistic model partitioning, cross-operation zero skipping, and model/config adaptive hardware reconfiguration. Splitwise \cite{patel2023splitwise} (section \ref{subsubsec:Splitwise}) improves LLM inference by separating workload onto different machines for high throughput, cost, or power efficiency. It allows for building both homogeneous and heterogeneous clusters depending on the optimization goal.

\subsection{Training Optimization: Challenges and Key Findings}
\label{training_optimizaiton_challenges_findings}

In the previous sections we have offered a comprehensive overview of training optimization (Section \ref{sec:training_optimization}) which includes model optimization (Section \ref{subsec:model_optimization}), size reduction optimization (Section \ref{subsec:size_reduction_optimization}), distributed training (Section \ref{subsec:distributed_training}), and heterogeneous training (Section \ref{subsec:heterogeneous_training}). In this section and the following paragraphs, we will discuss training optimization's challenges and key findings.

\emph{Challenges of Model Optimization:}
\begin{itemize}
    \item Resource constraints: LMs demand significant memory and computational power, limiting training and deployment on single devices.
    \item Balancing efficiency and accuracy: Optimizing LLMs requires finding a balance between efficient resource utilization and maintaining model performance.
    \item Memory bottlenecks: Distributing LMs across devices introduces memory limitations on each device.
    \item Communication overhead: Data exchange between devices during training can become a bottleneck, slowing down the process.
    \item Hardware heterogeneity: Efficiently utilizing devices with varying memory capacities and processing speeds in a distributed setting is challenging.
    \item Scalability limitation: Traditional methods might not scale well with increasing device numbers due to memory and communication constraints.

\end{itemize}

\emph{Key Findings:}
\begin{itemize}
    \item Algorithmic: Techniques like FlexGen \cite{b16}, LightSeq \cite{b58}, and NLP-Fast \cite{b55} improve efficiency by optimizing computations, memory access, and utilizing specialized hardware kernels.
    \item Model partitioning: Techniques like GPipe \cite{b3} and Megatron-LM \cite{b18} partition models for efficient processing across multiple devices.
    \item Fine-tuning for efficiency: Techniques like AlphaTuning \cite{b64} and LoRA \cite{hu2021lora} enable fine-tuning large models on limited memory by reducing the number of parameters requiring adjustment.
    \item Scheduler optimization: Techniques like TurboTransformers \cite{b11} improve response throughput and task execution on GPUs.
    \item Size reduction optimization: This approach focuses on reducing model complexity through techniques like quantization (reducing storage bits) and pruning (removing non-essential parts).
    \item Parallelism strategies: 1) Data parallelism: Distributes training data across devices for faster training. 2) Model parallelism: Splits the model across devices for parallel computations (tensor, pipeline, sequence parallelism). 3) Combined parallelism: Combines data and model parallelism for even faster training (PTD-P, ZeRO \cite{b17}, GPipe \cite{b3}).
    \item Memory optimization: ZeRO \cite{b17} optimizes memory for trillions of parameters, Activation Partitioning deals with activation memory efficiently, and ZeRO-Offload \cite{b19} and ZeRO-Infinity \cite{b1}, which allow training on single GPUs or limited resources by utilizing CPU and NVMe memory.
    \item Heterogeneous optimization: SWARM Parallelism \cite{b10} tackles unreliable devices with limited bandwidth by adapting workloads, NLP-Fast \cite{b55} optimizes execution on mixed platforms by pinpointing resource-heavy operations, and Splitwise \cite{patel2023splitwise} distributes work across heterogeneous machines considering different goals like throughput, cost, and power consumption.
    \item Automatic parallelism: Alpa \cite{b73} automatically generates execution plans for distributed model parallel training, applicable to diverse models.

\end{itemize}

Overcoming these challenges and leveraging these techniques, model training can be made more efficient, scalable, and accessible, paving the way for even more powerful and versatile LLMs.

\Figure[t!](topskip=0pt, botskip=0pt, midskip=0pt)[width=0.9\linewidth]{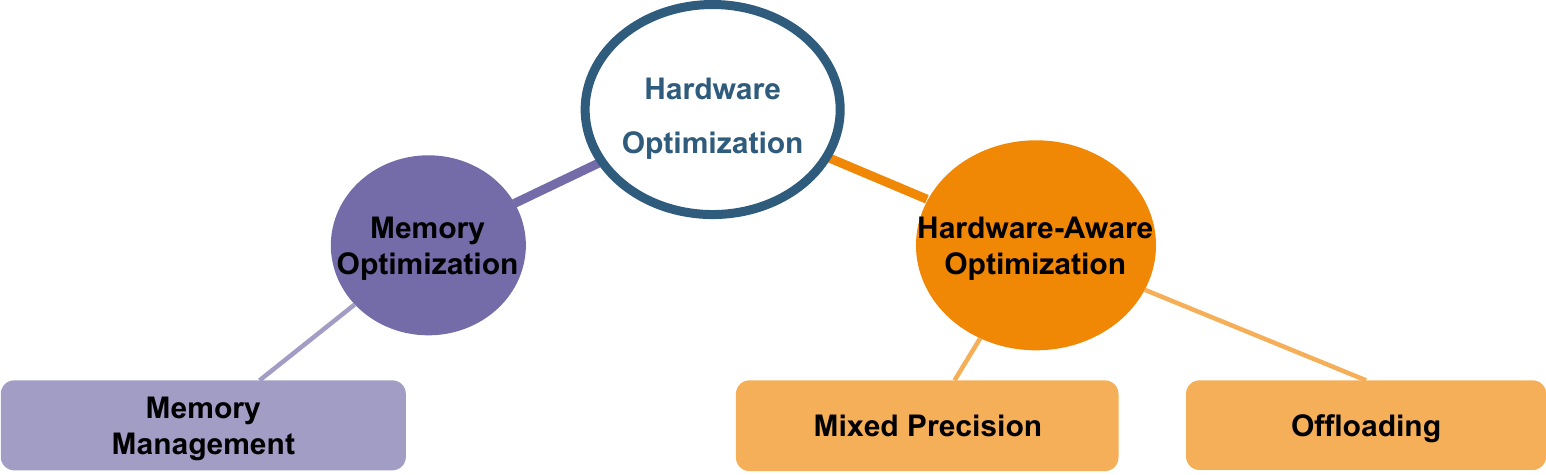}
{\textbf{Hardware optimization}\label{fig6}}

\begin{table*}[t]
\caption{\textbf{Comparative analysis between different strategies [A]}}
\label{rosta.t6}
\scalebox{0.88} 
{
\begin{tabular}{cp{175pt}p{135pt}p{130pt}}
Technique & Performance & Cost & Scalability \\ \hline   
 FlexGen \cite{b16} & 1) Batch size of 64 or 2048 tokens: Throughput speedup of 40$\times$ with a latency of 5000 seconds.2) Batch size of 256 or 8192 tokens: Throughput speedup of 79$\times$ with a latency of 12000 seconds.3) Batch size of 144 or 4608 tokens: Throughput speedup of 100$\times$ with a latency of 4000 seconds (using 4-bit quantization compression).
& Enabling high-throughput LLM inference on resource-constrained devices. Minimizing the need for multiple high-end GPUs. & Efficiently running the OPT-175B model on NVIDIA T4 (16 GB) GPUs, showcasing its ability to handle LMs on resource-constrained hardware. \\ \hline

SwapAdvisor \cite{b61} & Memory allocation: achieves up to 4$\times$ reduction in serving latency and boosts training throughput by 20\% to 1100\%.& Train models up to 12$\times$ beyond GPU memory limit. &  Supports efficient training and inference of LMs on standard GPUs, significantly extending their capability.  \\ \hline   

NLP-Fast \cite{b55} & Up to 2.92$\times$, 1.59$\times$, and 4.47$\times$ higher throughput on CPU, GPU, and FPGA respectively. & Reduce the need for high-end, resource-intensive hardware. & Scales to different hardware platforms (CPU, GPU, FPGA). \\ \hline
        
Byte Transformer \cite{b4} & Up to 87\% better performance compared to other frameworks (PyTorch JIT). & Reducing redundant computations. Improves the efficiency of running inference on transformer models, potentially reducing the cost of deployment. & Scales to different sequence lengths and transformer architectures. \\ \hline
        
Sheared LLAMA \cite{b62} & Superior performance compared to other open-source models of similar size. & 
Reduce the size of the LLaMA2-7B model to 1.3B and 2.7B parameters. Significantly reduced training cost (3\% of original). & Potentially efficient for deployment on resource-limited devices. \\ \hline
        
GrowLength \cite{b63} & Lower loss compared to fixed-length training (LLM128). & Potentially lower training cost (faster training). & Dynamically increasing training sentence length from 128 to 1024 tokens, enhancing efficiency in handling diverse text data, while maintaining lower loss.  \\ \hline
        
PagedAttention \cite{kwon2023efficient} & PagedAttention and vLLM achieve 2-4$\times$ higher throughput in LLM serving compared to existing systems, especially for large models, long sequences, and complex decoding algorithms. & vLLM's efficiency improvements can potentially reduce deployment costs by requiring fewer servers for the same workload. &  Handles large LLMs and vLLM's memory management scales well with diverse LLM architectures. \\ \hline
        
LightSeq \cite{b58} & Up to 14x and 1.4x speedups compared to TensorFlow and FasterTransformer respectively. & Reduced operational costs during deployment due to lower computational demands (inference on resource-constrained devices). & Well-suited for various transformer architectures. \\ \hline
        
LigthSeq2 \cite{b52} & Achieves 1.4-3.5$\times$ faster training compared to previous systems across various models and benchmarks, and 308\% speedup on WMT14 English-German machine translation compared to PyTorch. & Potentially lower cost (by enabling faster training). & Supports various transformer architectures, including BERT, GPT, and vision transformers.  \\ \hline
        
        
GPipe \cite{b3} & With a 557-million-parameter AmoebaNet model achieving 84.4\% top-1 accuracy on ImageNet-12 dataset. & Potentially lower cost (reduced hardware requirements). & Handling large and complex models across multiple accelerators, and achieving better quality than all bilingual models. \\ \hline
        
Megatron-LM \cite{b18} & Achieves SOTA results on NLP tasks (perplexity of 10.8 on WikiText103, 66.5\% accuracy on LAMBADA). & Allows utilizing fewer training instances or smaller model sizes for achieving similar performance. & Scales to train models with billions of parameters using multiple GPUs (demonstrated with 8.3B parameter model on 512 V100 GPUs). \\ \hline
        
AlphaTuning \cite{b64} & Maintains competitive performance on various tasks with over 1000$\times$ fewer parameters compared to full fine-tuning. & Reduces deployment costs by enabling less powerful hardware for inference. & Works with a wider range of LLMs, and its efficiency increases with even larger models due to quantization. \\ \hline
        
QFT \cite{li2023qft} & Maintains similar performance across various benchmarks. & Potentially reduces deployment costs due to lower memory requirements. Allows utilizing less powerful hardware, potentially leading to lower acquisition and maintenance costs. &  Handles large models with efficient memory management. Demonstrates successful fine-tuning of a 7B parameter LLaMA model, suggesting scalability for working with large language models.  \\ \hline
        
LOMO \cite{lv2023full} & Enables full parameter fine-tuning (65 billion parameter) of LLMs on limited resource GPUs. & Potentially reducing deployment costs through lower hardware acquisition and maintenance expenses. &  Especially suited for handling very large models. \\ \hline
        
AdaLomo \cite{lv2023adalomo} & Achieves scores of 30.8, 39.7, 51.0, and 56.9 on the LLaMA benchmark for models with 7B, 13B, 30B, and 65B parameters (performance comparable to AdamW) & Potentially lower training cost (due to reduced memory requirements). & Enables LLM training on resource-constrained environments by significantly reducing memory footprint while achieving comparable performance to AdamW, especially for models with a large number of parameters.\\ \hline
        
LoRA \cite{hu2021lora} & Maintains competitive performance on various tasks compared to full fine-tuning despite significantly fewer parameters (reduction by 10,000$\times$ for GPT-3) & Reduces deployment and training costs (3$\times$ reduction for GPT-3). & Applicable to various Transformer models (e.g., RoBERTa, DeBERTa, GPT-2, GPT-3). Designed to be efficient even for extremely LMs like GPT-3 (175B parameters). \\ \hline
        
TurboTransformers \cite{b11} & Enhances latency and throughput for transformer models, achieving better speed than PyTorch and comparable performance to TensorFlow-XLA, TensorRT, and FasterTransformers. & Reduces operational costs by optimizing memory usage through a sequence-length-aware memory allocation algorithm, ensuring efficient resource utilization. & Highly scalable, efficiently handling variable-length inputs with a sequence-length-aware batch scheduler to maximize throughput in diverse deployment scenarios. \\ \hline

PetS \cite{b56} & Enhances serving throughput by 1.53x on Desktop GPUs and 1.63x on Server GPUs. & Reduces the cost by requiring less storage space due to their smaller size compared to traditional transformers. & Supports up to 26$\times$ more concurrent tasks compared to existing serving systems. \\ \hline
        
QMoE \cite{frantar2023qmoe} & compresses a 1.6 trillion-parameter SwitchTransformer model to 160 GB (0.8 bits per parameter), resulting in a 20x reduction in size with minimal accuracy loss. & Achieves a 20$\times$ compression ratio, QMoE significantly reduces storage requirements. & Enables running trillion-parameter models on readily available hardware (e.g., single server with 4$\times$ NVIDIA A6000 GPUs) due to its compressed size. \\ \hline
        
SWARM Parallelism \cite{b10} & Trains large models on unreliable, heterogeneous devices with low network bandwidth. & Enables training on preemptible cloud instances or pooled resources from various regions, potentially reducing training costs compared to dedicated high-performance computing clusters. & It is designed for heterogeneous and unreliable devices, making it scalable to large deployments with varying computational power and network connectivity. \\ \hline

\end{tabular}
}
\end{table*}

\begin{table*}[t]
\caption{\textbf{Comparative analysis between different strategies [B]}}
\label{rosta.t7}
\scalebox{0.9} 
{
\begin{tabular}{p{60pt}p{190pt}p{150pt}p{115pt}}
Technique & Performance & Cost & Scalability \\ \hline   

GPTQ \cite{b66} & Achieves high accuracy with post-training quantization to 3 or 4 bits per parameter. & Reduces the bit width of the weights (down to 3 or 4 bits), significantly reducing the model size. & Allows inference of large GPT models on a single GPU due to its compressed size. \\ \hline

FPTQ \cite{li2023fptq} & Achieves SOTA performance on popular LLMs (BLOOM, LLaMA) with 4-bit weights and 8-bit activations (W4A8). & Utilizes a 4-bit weight quantization strategy, reduces the model size compared to full precision models. & Enables deployment of LLMs on resource-constrained devices by achieving high-performance W4A8 quantization (low memory footprint) without sacrificing accuracy.\\ \hline
        
Norm Tweaking \cite{b53} & Achieves high accuracy for large language models (GLM-130B, OPT-66B) even at 2-bit quantization. & Enables effective quantization down to even 2 bits, significantly reduces the model size compared to full precision. & Allows for deploying LLMs on devices with limited memory or computational power. \\ \hline
        
FineQuant \cite{kim2023finequant} & Up to 3.65× higher throughput on LLM inference with minimal accuracy loss for large models (OPT-175B). &  Focuses on weight-only quantization, reduces model size efficiently, potentially enabling deployment on less powerful hardware. &  Enables deployment of massive LLMs (like OPT-175B) on resource-constrained environments by achieving efficient weight-only quantization with high throughput and minimal accuracy loss. \\ \hline
        
PETALS \cite{b57} & Achieves faster inference for large language models, with an optimal setup inferring a 176 billion parameter model in 5.5 seconds. & 8-bit compression reduces resource requirements. & Scales by distributing computations across a network, enabling it to handle even larger models or more inference requests simultaneously. \\ \hline
        
        
QuantEase \cite{behdin2023quantease} & Up to 15\% better accuracy (perplexity, zero-shot) than GPTQ. Sub-3-bit quantization with minimal accuracy loss. & Enables effective quantization down to 3-bit or even lower precisions, significantly reducing the model size. & Quantizes large models (Falcon-180B) on 1 GPU in 3 hours. \\ \hline
        
LLM-Pruner \cite{ma2023llm} & Up to 95\% performance retention with 20\% parameter reduction (LLaMA, Vicuna, ChatGLM). & Potentially lower training cost due to less data needed for fine-tuning (3 hours, 50K samples). & Applicable to various LLM architectures (LLaMA, Vicuna, ChatGLM). \\ \hline
        
SparseGPT \cite{b67} & Up to 50\% sparsity (weight reduction) with minimal accuracy loss (perplexity). Larger models prune more easily (with less accuracy drop). & Potentially lower computational cost due to single-shot pruning (no retraining). & Processes very large models (OPT-175B, BLOOM-176B) efficiently. \\ \hline
        
        
Cramming \cite{b15} & Achieves reasonable performance by training on a single GPU in one day (trade-off between model size and training time). & Lower computational cost due to single GPU training. & Not designed for large-scale training, but explores trade-offs for resource-constrained settings. \\  \hline 

DFX \cite{b70} & 5.58x speedup in text generation compared to 4x NVIDIA V100 GPUs. & 8.21x more cost-effective than a GPU appliance with similar performance. & Designed for model parallelism across multiple FPGAs (scalability not explicitly quantified). \\ \hline

Narayanan \textit{et al}., \cite{b68} & High speed via pipeline, tensor, and data parallelism. & Potentially cost-efficient due to parallel processing, but not explicitly quantified. & Megatron-LM enables training LLMs (like trillion-parameter models) on thousands of GPUs by combining data, pipeline, and tensor parallelism (PTD-P) for efficient scaling and achieving high throughput.\\ \hline
 
ZeRO \cite{b17} & 10x speedup, trains trillion parameter models (8x larger than previous models). & Potentially reduces memory requirements for training large models. & Scales to trillion parameter models on large GPU clusters. \\ \hline
Colossal-ai \cite{b14} & Up to 2.76x faster training with various parallelism methods. & Potentially lower cost due to faster training and improved hardware utilization. & Modular design for customization and supports distributed training. \\ \hline
 
FlexFlow \cite{b72} & Achieves 2-10x speedup in performance for CNN workloads compared to existing architectures. & Improves power efficiency by 2.5-10x compared to existing architectures. & Highly scalable with growing computing engine size. \\ \hline
 
Alpa \cite{b73} & Matches Megatron-LM on GPT models, surpasses DeepSpeed on GShared MoE models (up to 9.7x speedup). & Alpa automates efficient model-parallel training for large deep learning models, potentially reducing development and infrastructure costs. & Designed for distributed deep learning. \\ \hline

ZeRO-Offload \cite{b19} & Trains 10x larger models on single GPUs (40 TFlops/GPU for 10B parameters). Supports models over 70B parameters with model parallelism. & Potentially lower training cost due to efficient single-GPU or smaller system training. & Enables large model training on single GPUs and scales to larger systems using model parallelism. \\ \hline

ZeRO-Infinity \cite{b1} & Trains models with trillions of parameters on GPU clusters. Enables fine-tuning on a single DGX-2 node. Achieves over 25 petaflops (exceeds peak performance by 40\%). & Potentially reduces memory requirements for training large models. & Highly scalable for training models with trillions of parameters. \\ \hline

Splitwise \cite{patel2023splitwise} & Up to 1.4x higher throughput for LLM inference compared to existing methods. & Potentially lower cost due to 20\% reduction in resource requirements for inference. & Scales well using homogeneous or heterogeneous machines for prompt computation and token generation phases. \\ \hline

Easy and Efficient Transformer (EET) \cite{b59} & Up to 27.43x speedup in transformer inference compared to Fairseq on A100 GPUs. Significant speedups over LightSeq and Faster Transformer as well. & Potentially lower cost due to reduced inference time (potentially leading to lower resource usage). & The library is designed to work with large model sizes and potentially scales well on different hardware configurations (2080Ti and A100 GPUs are mentioned). \\ \hline
Eliseev and Mazur \cite{eliseev2023fast} & Achieves 2-3 tokens per second inference speed on consumer GPUs for large sparse MoE models (Mixtral-8x7B). & Enables running large MoE models on limited hardware (consumer GPUs and free-tier Google Colab) potentially reducing costs. & Enables running large MoEs language models (like Mixtral-8x7B) on resource-constrained hardware (consumer GPUs and even free-tier Google Colab) by leveraging MoE-specific optimizations. \\ \hline

FP8-LM \cite{peng2023fp8} & Achieves 75\% faster training speed and 39\% memory reduction compared to BF16 training for a GPT-175B model on H100 GPUs (outperforms NVIDIA's Transformer Engine by 37\%). & Significantly reduces training costs for large models due to faster training and lower memory usage. & Reduces memory usage by 39\% and speeds up training of LLMs (like GPT-175B) by 75\% through an innovative FP8 mixed-precision framework, enabling training on resource-constrained environments. \\ \hline
       
\end{tabular}
}
\end{table*}

\section{Hardware Optimization}
\label{sec:hardware_optimization}
Hardware optimization is a systematic approach to improving the performance, efficiency, and functionality of computer hardware. By identifying and addressing bottlenecks in hardware architecture \cite{b17},  software, and the operating system, hardware optimization can enhance overall speed, reduce power consumption, and improve the reliability of hardware components (Fig. \ref{fig6}). Splitwise \cite{patel2023splitwise} (Section \ref{subsubsec:Splitwise}) is a technique to optimize hardware utilization by separating the prompt computation and token generation phases onto different machines. This approach allows designing clusters optimized for cost, throughput, and power consumption. The model achieves up to 1.4$\times$ higher throughput at 20\% lower cost or 2.35$\times$ higher throughput with the same cost and power.

\subsection{Memory Optimization}
\label{subsec:memory_optimization}
In the process of training deep learning models, memory usage is primarily attributed to various factors, including model parameters, layer activations, gradients, and optimizer states, such as momentum and variances in the Adam algorithm \cite{b14, b17}. The terms ``model states'' \cite{b17} or ``model data'' \cite{b14} encompass model parameters, gradients, and optimizer states collectively, while ``residual states'' \cite{b17} or ``non-model data'' \cite{b14} refer to layer activations, temporary buffers, and unusable fragmented memory collectively.

In this section, we will explain the common and recent approaches that have been used for increasing training throughput and loading larger models into GPU memory while training deep learning models.

\subsubsection{Memory Management}
\label{subsubsec:memory_management}
TurboTransformers \cite{b11} (section \ref{subsubsec:turbotransformers}), proposed a sequence length aware algorithm for memory allocation to efficiently balance memory allocation and deallocation, this algorithm overcomes the problem of variability of input sentence. LightSeq2 \cite{b52} introduces an innovative memory management approach, specifically designed for the Transformer structure. This strategy efficiently reduces peak memory consumption and minimizes the need for frequent allocation and release calls. Notably, LightSeq2 stands out as the pioneer in accelerating the entire training process of Transformers. In real-time applications where response time is crucial, model parallelism and pipeline parallelism can introduce significant delays due to the extra communication overhead caused by splitting tensors or layers, even with technologies like NVLink and GPUDirect. EET \cite{b59} (section \ref{subsubsec:eet}) focuses on minimizing memory usage for loading large models in online services. The proposed solution involves dynamic memory management, specifically targeting the reduction of memory consumption for activation caches and operation result buffers, as weights and certain pre-allocated caches are inherently difficult to compress. They introduce a dynamic CUDA memory management mechanism specifically designed to reduce CUDA memory usage for the same model size, unlike the manual memory allocation required by FT. 

\subsection{Hardware-Aware Optimization}
\label{subsec:hardware_aware_optimization}
Hardware-aware optimization (HAO) is the process of optimizing the hardware utilization of deep learning models to achieve maximum performance on specific hardware platforms \cite{b74}. In this section, we will explain offloading and mixed precision optimization.

\subsubsection{Offloading}
\label{subsubsec:offloading}
FlexGen \cite{b16} (Section\ref{subsubsec:flexgen}) presents an offloading framework for LLMs that optimizes I/O efficiency and throughput by considering computation schedule, tensor placement, and computation delegation. It utilizes a linear programming-based search algorithm and unifies the placement of weights, activations, and the KV cache, enabling significantly larger batch sizes compared to existing methods.

ZeRO-Offload \cite{b19} model facilitates the training of large model heterogeneous on GPU + CPU systems, enabling the handling of models up to 10$\times$ larger on a single GPU without sacrificing efficiency by using a unique optimal offload strategy. Also, the design achieves a highly scalable multi-GPU configuration by integrating the offload strategy with ZeRO-powered data parallelism, enabling ZeRO-Offload to achieve nearly linear scalability, and smooth integration with model-parallel training. This combination allows for the training of even larger models than using ZeRO-Offload or model parallelism independently. Moreover, the model enhances CPU performance by introducing a high-performance Adam optimizer, achieving a 6$\times$ improvement over SOTA Adam implementations. It also employs a one-step delayed parameter update strategy to overlap GPU forward and backward passes with CPU parameter updates. Additionally, the model's size has increased by a factor of 10 compared to widely used frameworks such as PyTorch. To maintain computational efficiency, the model minimizes data traffic to and from the GPU, increases GPU memory utilization, and allows offloading data and computation to the CPU. On a single NVIDIA V100 GPU, the model can achieve 40 TFlops/GPU for 10 billion parameters, and it can scale up to 128 GPUs when available. The model also supports model parallelism, enabling training models with more than 70 billion parameters on a single DGX-2 box, resulting in a 4.5$\times$ increase in model size compared to employing model parallelism alone.

Eliseev and Mazur \cite{eliseev2023fast} propose a model to efficiently run large sparse MoE language models on hardware with limited GPU memory. Using parameter offloading and leveraging the properties of MoE models enabled Mixtral-8x7B with mixed quantization to operate on desktop hardware and free-tier Google Colab instances. The study showed that some experts are reused between adjacent tokens, and early layers can predict subsequent experts. This led to an MoE-specific offloading strategy employing an LRU (Least Recently Used) cache and advanced prediction of needed experts. The model significantly improves speed, achieving 2-3 tokens per second on various consumer GPUs, and offers a practical solution for running large MoE models on limited hardware.

\subsubsection{Mixed Precision}
\label{subsubsec:mixed_precision}
Mixed precision training \cite{b75} proposes a method for training deep neural networks using half-precision floating-point numbers, aiming to reduce memory requirements by almost half and accelerate arithmetic on modern GPUs without compromising model accuracy or requiring adjustments to hyperparameters.

Cramming \cite{b15} conducts all experiments and ablation studies using a consistent setup that employs automated mixed precision for both standard 16-bit and 32-bit floating-point precision.

LightSeq2 \cite{b52} (section \ref{subsubsec:lightseq2}) optimizes the training process by implementing batched updates on reduced-precision parameters instead of numerous individual updates on full-precision parameters. In mixed precision training, where parameters and gradients are in FP16 during forward and backward propagation, maintaining FP32 copies is necessary for accuracy during the update values calculation. Typically, a system copies each piece of gradients, parameters to/from its FP32 counterpart in one training step, ensuring the accurate update of FP32 parameters with the loaded FP32 gradient by the trainer kernel.

FP8-LM \cite{peng2023fp8} introduces a novel FP8 automatic mixed-precision framework for training LLMs, optimizing mixed-precision and distributed parallel training through three levels of FP8 utilization. By gradually incorporating 8-bit gradients, optimizer states, and distributed learning, the framework significantly enhances training efficiency. During the training of a GPT-175B model on the H100 GPU platform, the FP8 framework reduced memory usage by 39\% and increased training speed by 75\% compared to the BF16 framework, outperforming Nvidia's Transformer Engine by 37\%. This advancement leads to substantial cost reductions for training large models and is adaptable to various tasks such as instruction tuning and reinforcement learning with human feedback.

\subsection{Hardware Optimization: Challenges and Key Findings}
\label{hardware_optimizaiton_challenges_findings}

\emph{Challenges of Hardware Optimization:}
\begin{itemize}
    \item Memory limitation: Deep learning models can require vast amounts of memory to store parameters, activations, and gradients. This limits the size and complexity of models that can be trained on a single device.
    \item Limited hardware utilization: Traditional training methods may not fully utilize the capabilities of modern hardware like GPUs.
    \item Balancing speed and accuracy: Techniques like mixed precision training aim to improve training speed by reducing memory usage, but this can potentially compromise model accuracy.

\end{itemize}

\emph{Key Findings:}
\begin{itemize}
    \item Memory management: Techniques like sequence length aware allocation and dynamic memory management can significantly reduce memory usage during training.
    \item Hardware-aware optimization: Offloading computations to CPUs or leveraging mixed precision training can improve hardware utilization and training speed.
    \item Model parallelism: Splitting models across multiple devices can handle larger models but can introduce communication overhead, impacting training speed.
    \item Large model training: Frameworks like ZeRO-Offload \cite{b19} enable training models significantly larger than what a single GPU can handle.
\end{itemize}

In the domain of hardware optimization, a continuous stream of novel methodologies is emerging, demonstrably expanding the frontiers of feasibility within the training paradigm.

\Figure[t!](topskip=0pt, botskip=0pt, midskip=0pt)[width=0.9\linewidth]{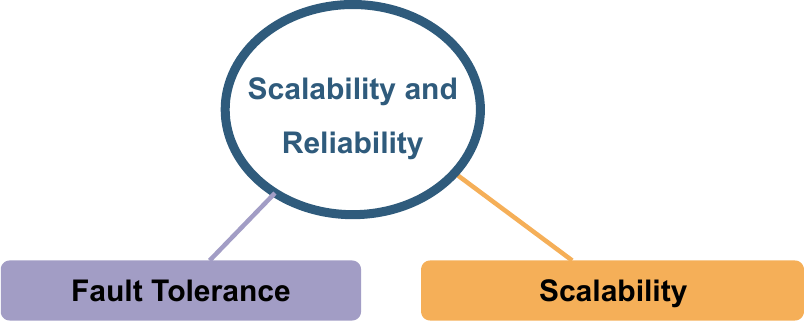}
{\textbf{Scalability and reliability optimization}\label{fig7}}

\section{Scalability and Reliability Optimization}
\label{sec:scalability_and_reliability}
Scalability optimization focuses on improving hardware systems' capacity to flexibly handle varying workloads, enabling smooth scaling adjustments to meet evolving demands\cite{b1, b5, b17, b18, b19, b68}, and reliability optimization aims to strengthen the dependability and stability of hardware infrastructure, reducing the likelihood of failures, errors, or disruptions \cite{b10, b57} (Fig. \ref{fig7}).

\subsection{Fault Tolerance}
\label{subsec:fault_tolerance}
SWARM Parallelism\cite{b10} (section \ref{subsec:heterogeneous_training}) allows high-performance devices to handle inputs from several preceding sources, share their outcomes with less powerful peers, and adjust the workload distribution in the event of a failure, enhancing resource utilization. The model ensures continuous training and boosts overall efficiency by redistributing workload in case of device failure or premature termination. 

PETALS \cite{b57} (section \ref{subsubsec:petals}) is a distributed Transformer model that can be easily scaled and fault-tolerant. It uses a load-balancing algorithm to distribute servers evenly among Transformer blocks and a routing algorithm to find the fastest path for inference. It also stores past inputs to each server in case one fails, so that the client can quickly continue with a replacement server. PETALS is a reliable and scalable Transformer model that can be used for both inference and training. It uses a combination of load balancing, routing, and fault tolerance to ensure that it can handle network disruptions and server failures without impacting performance.

\subsection{Scalability}
\label{subsec:scalability}
ZeRO-Offload \cite{b19} is a highly scalable multi-GPU design achieved through an integrated offload strategy and ZeRO-powered data parallelism. This combination leads to nearly linear scalability, allowing for the training of significantly larger models than when using ZeRO-Offload or model parallelism independently. The model further optimizes CPU execution with a high-performance Adam optimizer, resulting in a 6 time higher than SOTA Adam implementation. Despite a growth in model size by a factor of 10, the approach minimizes data traffic to and from the GPU, maximizes GPU memory utilization, and facilitates offloading data and computation to the CPU. ZeRO-Offload maintains a single copy of optimizer states in CPU memory, ensuring constant communication volume and CPU computation, regardless of data parallelism. This design choice enables excellent scalability on up to 128 GPUs, and ZeRO-Offload can also be combined with model parallelism for higher memory savings when multiple GPUs are available.

\subsection{Scalability and Reliability Optimization: Challenges and Key Findings}
\label{scalability_and_reliabilty_optimizaiton_challenges_findings}

\emph{Challenges of Scalability and Reliability:}
\begin{itemize}
    \item In the context of optimizing LLMs, a trade-off exists between achieving high scalability and maintaining reliability. Scalability, which involves handling increased workloads, often necessitates the integration of more complex components. However, this added complexity can introduce new potential points of failure, thereby impacting the system's overall reliability. Balancing these two objectives is crucial to ensure both effective performance and robustness in large-scale deep learning systems.
   
\end{itemize}

\emph{Key Findings:}
\begin{itemize}
    \item Fault tolerance: This approach involves creating mechanisms to handle failures gracefully. Two notable techniques are SWARM Parallelism \cite{b10} and PETALS \cite{b57}. SWARM Parallelism distributes the workload across multiple devices and compensates for failures by redistributing tasks if a device fails. Similarly, PETALS, a distributed Transformer model, employs load balancing and routing strategies to maintain smooth operation even in the event of server failures.
    \item Scalability techniques: Technique like ZeRO-Offload \cite{b19} achieve high scalability for training large models. This method combines data parallelism with an offloading strategy, minimizing data traffic and maximizing resource utilization.
\end{itemize}

\begin{table*}[h!]

\caption{\textbf{Summary on reviewed papers excluding those already covered in Tables \ref{rosta.t3} and \ref{rosta.t4} or the main text.}}
\label{rosta.t8}
\scalebox{0.75} 
{
\begin{tabular}{p{90pt}p{240pt}p{310pt}}
\hline
Studies&
Aims&
Outcomes \\
\hline
ZeRO-Infinity \cite{b1} & Effectively breaks the GPU memory barrier, making large-scale model training \textbf{accessible on constrained resources}. & 
Models scaled to trillions of parameters on GPU clusters, with fine-tuning possible on a single NVIDIA DGX-2 node. Consistently reach over 25 petaflops, exceeding peak performance by 40\%.
\\
\hline
ZeRO \cite{b17} & Efficiently train large models, overcome limitations of the existing methods. & 
Achieved 15 Petaflops during training with 100B parameter models on 400 GPUs, showing super-linear speedup. This means an 8$\times$ larger model size and a 10$\times$ performance boost compared to prior benchmarks.
\\
\hline
SwapAdvisor \cite{b61} & An approach for deep learning memory management, enables training and serving of large models \textbf{despite limited GPU memory}. & Training models up to 12$\times$ beyond the usual GPU memory limit.
\\
\hline

Sheared LLAMA \cite{b62} & A \textbf{dynamic batch loading}, efficiently adjusts the composition of sampled data within each training batch based on varying losses observed across different domains. & The LLaMA2-7B model is reduced to 1.3B and 2.7B parameters, needing only 3\% of the usual computing resources for training. Tests were conducted using a maximum of 16 Nvidia A100 GPUs (80 GB).
\\
\hline

GrowLength \cite{b63} & Accelerates the pre-training process of LLMs by dynamically and \textbf{progressively growing the training sentence length}. & Three different setups were investigated: LLM128 with 128-token sentences and 0.36B parameters; LLM1024 with longer sentences but the same total tokens; and GrowLength, which grows from 128 to 1024 tokens. GrowLength shows lower loss than LLM128, emphasizing its computational efficiency and practicality in resource-limited scenarios.
\\
\hline

AlphaTuning \cite{b64} & It combines quantization of PLMs with fine-tuning, \textbf{only a subset of quantized parameters is fine-tuned for the target task}. & 
Applied to GPT-2 and OPT, achieved over 10x compression under 4-bit quantization and reduced trainable parameters by over 1,000-fold, maintaining competitive performance on various tasks.
\\
\hline

Cramming \cite{b15} & Investigates the trade-offs involved in \textbf{scaling down} language model and training on a single GPU in one day. & Explored two setups: one with a classical RTX2080Ti GPU and another with modern RTXA4000 or RTXA6000 GPUs, each paired with 4 CPU cores and 32 GB RAM.
\\
\hline

SWARM Parallelism \cite{b10} & Train a large model with \textbf{unreliable heterogeneous devices} with low network bandwidth by using dynamically generated, \textbf{randomized pipelines}. & Trained a large transformer language model with 1B shared parameters using compression strategy on preemptive T4 GPUs with network bandwidth below 200Mb/s.
\\
\hline

GPTQ \cite{b66} & High accurate, and efficient \textbf{post-training quantization} method which is known as a new one-shot weight quantization. & Precisely quantized models to 3 or 4 bits per parameter, taking a few hours on models with hundreds of billions of parameters. Experiments on OPT-175B and BLLOM-176B, it took around 4 GPU hours with minimal loss of accuracy compared to the uncompressed baseline.
\\
\hline

Norm Tweaking \cite{b53} & Presents a strategy to minimize computational and storage demands in large language models without compromising their performance. & Achieving high accuracy, GLM-130B and OPT-66B maintain accuracy even at 2-bit quantization. Improvements in weight-only and joint quantization surpass existing post-training quantization (PTQ) methods.
\\
\hline

SparseGPT \cite{b67} & A \textbf{post-training pruning} method to prune massive GPT-family models efficiently and accurately. & 
Ran on open-source models OPT-175B and BLOOM-176B in under 4.5 hours. Achieved 50-60\% unstructured sparsity with minimal impact on perplexity and removed over 100 billion weights with negligible accuracy loss.
\\
\hline

ZeRO-Offload \cite{b19} & Democratize large-scale model training, making it accessible to a wider audience. & Trained large model heterogeneously on GPU + CPU systems, achieving 10$\times$ greater model size on a single GPU without sacrificing efficiency. Achieved 40 TFlops/GPU for 10 billion parameters on a single NVIDIA V100 GPU, scalable up to 128 GPUs. Supports model parallelism, enabling training models with over 70 billion parameters on a single DGX-2 box, resulting in a 4.5$\times$ increase in model size.
\\
\hline

Alpa \cite{b73} & Automated system that generates execution plans for \textbf{distributed model-parallel training}. & Achieves comparable training performance to Megatron-LM on GPT models and surpasses DeepSpeed on GShared MoE models with up to 9.7$\times$ speedup.
\\
\hline

Efficient large-scale \par language model \par training on GPU clusters\par using megatron-LM \cite{b68} & Introduces PTD-P, a novel technique for training LLMs on GPU clusters, combining pipeline, tensor, and data parallelism for high computational performance and scalable training. & 
Offers significant performance enhancements over ZeRO-3, delivering a 70\% improvement for models with 175 and 530 billion parameters, mainly due to reduced cross-node communication overhead.
\\
\hline

DFX\cite{b70} & A low-latency multi-FPGA appliance for accelerating transformer-based text generation.& 
Utilized four Xilinx Alveo U280 FPGAs to evaluate performance on the GPT-2 language model, achieving a 5.58$\times$ speedup and 3.99$\times$ energy efficiency improvement over four NVIDIA V100 GPUs. Demonstrated to be 8.21$\times$ more cost-effective than a GPU appliance with similar performance.
\\
\hline

Colossal-AI \cite{b14} & A \textbf{unified deep learning system}. This system would streamline the process of \textbf{training complex models with billions of parameters on multi-GPU clusters}.& 
Colossal-AI is a user-friendly system that offers various parallel training techniques and integrates with advanced methods for enhanced performance. Notably, Colossal-AI achieved training speedups of up to 2.76 times for large models compared to traditional methods.
\\
\hline

LoRA \cite{hu2021lora} & A method that improves LLM adaptation by \textbf{reducing the number of trainable parameters during fine-tuning}.& 
Reducing trainable parameters by 10,000x and memory usage by 3$\times$ compared to traditional methods. It maintained or improved model performance while offering faster training and efficient task-switching.
\\
\hline

AdaLomo \cite{lv2023adalomo} & Aims to address the \textbf{memory limitations of existing optimizers} like AdamW by using memory-efficient techniques while retaining the benefits of adaptive learning rates. & The outcome is a successful optimizer that achieves performance comparable to AdamW on various tasks. This allows for training LLMs with significantly less memory usage, making large-scale LLM training more accessible.

\\
\hline

Li \textit{et al} (PagedAttention) \cite{kwon2023efficient} & A novel attention algorithm inspired by virtual memory. This technique aims to improve memory efficiency in LLM training by \textbf{reducing fragmentation and enabling efficient sharing.} &  Achieves significant improvements in throughput (2-4$\times$) for LLM serving, particularly for large models, complex decoding algorithms, and long sequences.

\\
\hline

FlexFlow \cite{b72} &  A novel dataflow architecture that leverages complementary parallelism effects to achieve \textbf{improved resource utilization within CNN accelerators.} &  The outcome is a significant improvement in performance (2-10$\times$ speedup) and power efficiency (2.5-10$\times$) compared to existing architectures on various CNN workloads.
\\
\hline

QFT \cite{li2023qft} &  Develop a memory-efficient framework (QFT) for \textbf{fine-tuning LLMs.} &  Achieves significant memory reduction during fine-tuning by utilizing quantization techniques, the Lion optimizer, and integer-based model states. This allows fine-tuning of large models on a single GPU with minimal performance loss compared to traditional methods.
\\
\hline

QMoE \cite{frantar2023qmoe} &  A framework that \textbf{significantly reduces memory usage for large MoE models}. & Achieves significant memory reduction through a custom compression algorithm that shrinks models to less than 1 bit per parameter, enabling execution on affordable hardware like single servers with multiple GPUs.
\\
\hline

FPTQ \cite{li2023fptq} &  A \textbf{post-training quantization} technique for compressing LLMs. & Achieves significant memory reduction and computational efficiency during inference with minimal accuracy loss.

\\
\hline

FineQuant \cite{kim2023finequant} &  A method to \textbf{improve the efficiency of LLM inference}. & Achieves significant memory reduction and faster inference speeds with minimal accuracy loss for LLMs.

\\
\hline

QuantEase \cite{behdin2023quantease} &  A framework for \textbf{efficiently deploying LLMs} by making them smaller.& Achieved SOTA  performance in quantizing LLMs. This also improved perplexity and zero-shot accuracy by up to 15\% compared to existing methods. It quantifies large models like Falcon-180B on a single GPU in 3 hours. 

\\
\hline

LLM-Pruner \cite{ma2023llm} & A task-agnostic framework for \textbf{compressing large LLMs }with minimal reliance on the original training data. & Compressed LLMs (LLaMA, Vicuna, ChatGLM) by 20\% while maintaining 94.97\% of their original performance.

\\
\hline

Li \textit{et al} \cite{b71} & A \textbf{ memory-efficient method }called sequence parallelism to train Transformers with much longer sequences on GPUs. & Achieved $\times$ longer maximum sequence length and 13.7 $\times$ larger batch size compared to SOTA tensor parallelism on 64 GPUs. With sparse attention, it can handle sequences over 27 $\times$ longer than existing methods.

\\
\hline

FP8-LM \cite{peng2023fp8} & A new framework called \textbf{FP8-LM for training LLMs using mixed precision} to improve efficiency. & Reduced memory usage by 39\% and increased training speed by 75\% compared to the BF16 framework. It outperformed Nvidia's Transformer Engine by 37\% in training speed.

\\
\hline

LOMO \cite{lv2023full} & A memory-efficient \textbf{ training method for LLMs on limited GPU resources}. & Enables fine-tuning of massive LLMs (65 billion parameters) on consumer-grade GPUs (RTX 3090) by significantly reducing memory usage compared to traditional methods.

\\
\hline

Eliseev and Mazur \cite{eliseev2023fast} & A method to run large, \textbf{sparse MoE LMs on limited GPU memory.} & This method uses parameter offloading and MoE properties to run on desktop hardware and free Google Colab instances. It leverages expert reuse and early layer prediction to achieve an MoE-specific offloading strategy. This significantly improves speed (2-3 tokens per second) on various consumer GPUs, making large MoE models practical for limited hardware.

\\
\hline

\multicolumn{3}{p{500pt}}{Bold text in the "Aims" column indicates the framework's primary area of specialization or the range of tasks it is designed to address. }\\
\multicolumn{3}{p{500pt}}{This table summarizes the reviewed papers excluding those already covered in Tables \ref{rosta.t3} and \ref{rosta.t4} or the main text. }\\

\end{tabular}
}
\end{table*}

\section{Case Studies}
\label{sec:case_studies}
The following case studies delve into the practical application of advanced optimization strategies on LLMs. With the rapid growth and increasing complexity of LLMs, efficient deployment and execution have become critical challenges. These case studies illustrate how cutting-edge techniques in model compression, pruning, and inference optimization can significantly enhance the performance and feasibility of deploying these massive models on more accessible hardware. By examining specific implementations and outcomes, these examples provide valuable insights into overcoming the computational and resource constraints associated with large-scale language models, thereby promoting their broader adoption and utility in real-world applications.

\subsection{Optimizing Model Training with SparseGPT}
\label{subsec:optimizing_model_training_with_sparsegpt}

\emph{Background:} LLMs like GPT-3 have billions of parameters, which pose significant challenges in terms of storage, computational requirements, and energy consumption. Pruning, or removing less important parameters, can help mitigate these issues, but traditional pruning methods often require multiple iterations of fine-tuning, which is computationally expensive. This approach (SparseGPT \cite{b67}) proposes a one-shot pruning method that significantly reduces the number of parameters without the need for extensive retraining.

\emph{Context and Problem:} In this case study, the focus is on training a LLM with billions of parameters on limited hardware. The initial challenge was the high computational and memory requirements that exceeded the capabilities of available resources, making it difficult to efficiently train the model within a reasonable timeframe and budget.

\emph{Optimization Strategy:} The primary optimization strategies involved in SparseGPT are: 

\emph{One-Shot Pruning:} To achieve significant sparsity in the LLM in a single pruning step, eliminating the need for iterative pruning and retraining. One-Shot Pruning: SparseGPT implements its pruning strategy through a streamlined process. First, a thorough  model analysis  is conducted to pinpoint parameters that can be removed without significant impact. This analysis leverages pruning criteria that assess parameter importance without requiring gradient calculations, saving on computational resources. Finally, SparseGPT employs a single step pruning approach, achieving substantial sparsity (at least 50\% for massive models) in a single step. This one-shot approach significantly reduces the time and complexity compared to iterative pruning methods.

\emph{Unstructured Sparsity:} To reduce the number of parameters while maintaining model accuracy through unstructured pruning, where individual weights are removed based on their importance. This approach focuses on eliminating individual weights within the model that are deemed less important. By analyzing the model's internal structure, SparseGPT achieves impressive sparsity levels of 50-60\%, significantly reducing model size. This aggressive pruning strategy is remarkable because it achieves this with minimal impact on the model's ability to perform language modeling tasks accurately. For instance, SparseGPT can remove over 100 billion weights from massive models like OPT-175B and BLOOM-176B without compromising their performance on language modeling tasks.

\emph{Parametrization without Gradient Dependence:} To leverage the parametrization of massive GPT models to enable pruning without relying on gradient information. This method allows the identification of sparse counterparts within a close range of the original dense model, ensuring these sparse models maintain similar performance. Interestingly, the strategy highlights that larger models are even easier to prune using this approach. They experience minimal accuracy drops even at significant sparsity levels (e.g., 50\%). This observation underscores the effectiveness of the parametrization technique in enabling aggressive pruning while preserving model performance.

\emph{Outcomes:} The application of SparseGPT led to remarkable results:
\begin{itemize}
    \item \textbf{Model size reduction:} SparseGPT achieved 50-60\% sparsity, significantly reducing the model size by removing more than 100 billion weights in models like OPT-175B and BLOOM-176B.
    \item \textbf{Processing time:} The pruning process was completed in less than 4.5 hours for the largest open-source models, demonstrating high efficiency.
    \item \textbf{Accuracy maintenance:} The pruned models exhibited negligible increases in perplexity and retained performance levels very similar to their dense counterparts.
    \item \textbf{Scalability:} The study revealed that larger models are easier to prune, with practically no accuracy decrease observed at 50\% sparsity.
\end{itemize}
This case study demonstrates the efficacy of SparseGPT's one-shot pruning approach for reducing the size of massive language models. By leveraging unstructured sparsity and parametrization strategies without gradient dependence, SparseGPT achieves substantial reductions in model size and resource requirements while maintaining high levels of performance. This approach enables more efficient and accessible deployment of large language models in various applications, making them more practical for real-world use.

\subsection{Enhancing Inference Efficiency with QMoE}
\label{subsec:enhancing_inference_efficiency_with_QMoE}
\emph{Background:} LLMs with trillions of parameters are becoming increasingly common. However, training and deploying these models is challenging due to their immense computational and memory demands. Existing compression techniques struggle to handle such large models effectively. QMoE \cite{frantar2023qmoe} framework addresses this challenge by introducing novel compression methods to make these models more practical for real-world use.

\emph{Strategy Selection:} QMoE was chosen as the optimization strategy. This approach allows for the compression of large models by quantizing their parameters to extremely low precision, which drastically reduces the model size while maintaining its performance. This strategy is particularly useful for handling the large parameter counts typical of MoE models.

\emph{Optimization Strategy:} The core optimization strategies involved in QMoE are: 

\emph{Scalable Compression Algorithm:} QMoE tackles the challenge of massive model sizes with a scalable compression algorithm. This innovative technique achieves impressive sub-1-bit compression for trillion-parameter MoE models, without requiring retraining. In the case of the SwitchTransformer-c2048 model, this translates to a dramatic size reduction from 3.2 TB to a mere 160 GB (roughly 0.8 bits per parameter). Remarkably, this is achieved with minimal compromise on accuracy, as measured by performance on pretraining validation tasks and zero-shot data.
    
\emph{Customized Compression Format and GPU Kernels:} QMoE takes advantage of custom designed GPU kernels to unlock the potential of its compressed format. These specialized kernels enable swift, on-the-fly decoding of the model, ensuring efficient processing during use. This allows the compressed model to run seamlessly on common hardware like 8 NVIDIA RTX 3090 or 4$\times$ NVIDIA A6000 GPUs. Even with this readily available hardware, the runtime overhead stays below 5\% compared to an uncompressed model, which would require a staggering 20 times more GPUs.

\emph{Outcomes:} The implementation of QMoE resulted in significant improvements:
\begin{itemize}
\item \textbf{Compression ratio:} The model size was reduced by approximately 95\%, allowing the SwitchTransformer-c2048 model to fit within the memory constraints of standard hardware. This reduction from 3.2 TB to less than 160 GB translates to a compression ratio of around 0.8 bits per parameter.
\item \textbf{Inference speed:} The QMoE framework enables the efficient execution of massive MoE models on commodity hardware with a runtime overhead of less than 5\%. This efficiency allows the trillion-parameter SwitchTransformer-c2048 model to run on a single commodity GPU server.
\item \textbf{Accuracy:} Despite the substantial compression, the model maintains high performance on pretraining validation tasks and zero-shot data, with only a minor decline in accuracy.
\end{itemize}
This case study demonstrates the feasibility of deploying trillion-parameter models in real-world applications through the use of advanced compression techniques. The QMoE approach not only reduces resource requirements but also enhances the deployability of cutting-edge language models across various environments. By leveraging a scalable compression algorithm, a customized compression format, and bespoke GPU kernels, QMoE achieves significant improvements in model efficiency and performance. This makes large-scale models more accessible and practical for real-world applications. It addresses key limitations of MoE architectures and promotes their wider adoption, paving the way for further research and advancements in this field.

\section{Discussion}
\label{discussion}
This section examines optimization and acceleration techniques for LLMs. We will discuss the relevant libraries and frameworks that facilitate these advancements, alongside challenges and key findings of various optimization strategies.

\subsection{LLM Training Challenges}
\label{llm_training_chalelnges}
Training LLMs poses significant challenges due to their complexity and resource requirements. Recent advancements in frameworks like GPipe \cite{b3}, ByteTransformer \cite{b4}, Megatron-LM \cite{b18}, LightSeq2 \cite{b52}, and CoLLiE \cite{lv2023collie} have made significant strides in addressing these challenges:

\emph{Distributed training:} As LLMs become increasingly complex, training them on a single device becomes impractical. Megatron-LM \cite{b18} and CoLLiE \cite{lv2023collie} address this by employing distributed training algorithms that partition the model across multiple GPUs. This approach enables parallel processing and significantly accelerates training times. By distributing the workload, these frameworks mitigate the memory bottlenecks that arise when trying to train massive models on single devices.

\emph{Efficiency and speed:} Efficiency and speed are critical for the practical deployment of LLMs. LightSeq2 \cite{b52} enhances training speed through system-level optimizations such as layer-specific kernels and mixed-precision training, which improve GPU utilization and reduce memory usage. Similarly, ByteTransformer \cite{b4} is designed to accelerate transformer models, particularly for variable-length inputs in NLP tasks, thereby improving performance and reducing latency.

\emph{Memory Management:} Efficient memory allocation is crucial for training large models. CoLLiE \cite{lv2023collie}  addresses memory constraints in LLM training through a comprehensive strategy. It implements 3D parallelism to effectively distribute memory across training machines and GPUs. This approach allows CoLLiE to train large language models even in environments with limited resources.

\emph{Fine-Tuning and Performance:} CoLLiE \cite{lv2023collie} also focuses on enhancing specific capabilities of LLMs through PEFT methods. These methods allow models to be fine-tuned for particular tasks or user instructions without compromising their overall performance. This targeted improvement is vital for developing models that can adapt to specific application needs while maintaining high general performance.

\subsection{LLM Training Key Findings}
\label{llm_training_key_findings}
The advancements in these frameworks have led to several significant findings:

\emph{GPipe:} Demonstrates the successful training of a large multilingual transformer model, achieving superior results compared to smaller, individually trained models \cite{b3}.

\emph{ByteTransformer:} Outperforms existing frameworks in terms of performance for BERT-like transformers on various benchmarks \cite{b4}.

\emph{Megatron-LM:} Enabled the training of LLMs with billions of parameters, achieving SOTA results on numerous NLP tasks while providing high throughput \cite{b18}.

\emph{LightSeq2:} Accelerates transformer model training by up to 308\%, showcasing substantial performance improvements \cite{b52}.

\emph{CoLLiE:} Introduces collaborative training methodologies that improved efficiency and effectiveness in training large models like LLaMA-65B, exploring ways to enhance specific functionalities without impacting overall performance \cite{lv2023collie}.

\subsection{LLM Inference Challenges}
\label{llm_tinference_chalelnges}
Efficient inference of LLMs is critical for their practical application, as these models are computationally expensive due to their size and complexity. In this section, we will discuss and explore the challenges and key findings of various frameworks and libraries designed to enhance the efficiency of LLM inference.

\emph{Computational expense}: The massive size and complex architecture of LLMs make traditional inference methods inefficient, especially on resource-constrained devices.

\emph{Balancing speed, accuracy, and resource utilization:} Achieving an optimal balance between these factors are crucial for real-world deployment of LLMs.

\subsection{LLM Inference Key Findings}
\label{llm_inference_key_findings}
\emph{Hardware specialization:} Frameworks like Splitwise \cite{patel2023splitwise} improve inference by separating compute-intensive and memory-intensive phases onto different machines with specialized hardware. This targeted approach optimizes resource usage and enhances performance.

\emph{Resource optimization:} FlexGen \cite{b16} employs techniques such as I/O scheduling, compression, and distributed processing to efficiently utilize resources across CPUs, GPUs, and disk storage. This holistic resource management approach significantly improves inference efficiency.

\emph{Algorithmic optimizations:}
Libraries like EET \cite{b59} and LightSeq \cite{b58} implement custom algorithms and advanced memory management techniques to accelerate inference on GPUs. These optimizations reduce latency and improve throughput, making LLM inference more practical for real-time applications.

\emph{Heterogeneous platforms}
NLP-Fast \cite{b55} leverages different hardware platforms, including CPUs, GPUs, and FPGAs, by identifying performance-critical operations and applying targeted optimizations. This flexibility allows for efficient inference across various hardware configurations.

\emph{Distributed Inference}
PETALS \cite{b57} facilitates collaborative inference and fine-tuning of LLMs across a network, enabling scalable and efficient resource utilization. This approach allows for distributed processing, which is essential for handling large-scale inference tasks.

\subsection{LLM Deployment and Serving Challenges}
\label{llm_deployment_chalelnges}
Deploying and serving LLMs in real-world applications presents several challenges. This section explores these challenges, key findings from recent advancements, and future directions for making LLM deployment and serving more efficient and accessible.

\emph{Memory limitation}: LLMs often exceed the memory capacity of a single GPU, complicating their deployment and serving in practical applications.

\emph{Scalability:} Handling multiple user requests simultaneously requires efficient scaling solutions to manage the large and complex models effectively.

\emph{Variability of input:} LLM performance can be inconsistent when dealing with input sequences of varying lengths, necessitating dynamic memory allocation strategies to maintain efficiency.

\emph{Ease of deployment:} Integrating complex LLM serving systems into existing workflows can be challenging, particularly for researchers and practitioners without extensive expertise in the field.

\subsection{LLM Deployment and Serving Key Findings}
\label{llm_deployment_key_findings}

\emph{PagedAttention (vLLM) :} This algorithm breaks down the KV cache into manageable blocks, minimizing wasted memory and enabling efficient sharing across requests. This is a significant improvement for processing large LLMs \cite{kwon2023efficient}.

\emph{Efficient GPU utilization (TurboTransformers):} Utilizes techniques like parallel GPU kernels and dynamic batch scheduling to optimize performance on GPUs, resulting in faster inference for transformer-based models  \cite{b11}.

\emph{System-level optimizations (LightSeq2):} Demonstrates how system-level optimizations within the training process can significantly improve training speed and efficiency, translating to faster deployment of LLMs \cite{b52}.

\subsection{Hardware Optimization in LLM}
\label{hardware_optimization_in_llm}
Optimizing hardware for LLM involves overcoming memory limitations and improving utilization. Key findings include efficient memory management, hardware-aware optimization, and model parallelism. Future research should focus on efficient offloading strategies and advanced mixed precision training.

\subsection{Scalability and Reliability Optimization in Hardware Systems}
\label{scalability_and_reliability_optimizaitn_in_hardware_system}
Achieving scalable and reliable hardware systems requires balancing complexity with reliability. Techniques like SWARM parallelism and ZeRO-Offload \cite{b19} improve fault tolerance and scalability. Future research should develop advanced fault tolerance mechanisms and optimize for new hardware.

These advancements collectively enhance the efficiency, scalability, and accessibility of LLM training, inference, deployment, and serving, paving the way for more powerful language models.

\section{Conclusion and Future Directions}
\label{conclusion_and_future_directions}
This SLR investigated optimization and acceleration techniques for LLMs. We identified the challenges associated with training, inference, and system serving for LLM with billion or trillion parameters. We presented a structured taxonomy of optimization techniques alongside a comprehensive analysis of recent libraries and frameworks. Following the PRISMA statement, we meticulously analyzed 65 relevant studies published between 2017 and December 2023.
Our proposed taxonomy provides a roadmap for researchers to navigate the diverse landscape of optimization strategies and select the most suitable approaches for their specific tasks. Additionally, the review of libraries and frameworks empowers researchers to efficiently train and deploy LLMs, accelerating progress in real-world applications. Furthermore, the inclusion of two in-depth case studies demonstrates practical approaches to optimizing model training and enhancing inference efficiency, highlighting how resource limitations can be addressed while maintaining performance.

While recent advancements in LLM frameworks and optimization techniques are promising, further research is crucial to unlock their full potential. We identified several key areas for future exploration, focusing on enhanced efficiency, scalability, and flexibility for LLMs.

\subsection{Optimization for Resource-Constrained Environments}
\label{optimization_for_resource_constrained_enviroment}

\emph{Hybrid processing:} Develop hybrid processing techniques, where computation is split between GPUs and CPUs to optimize memory usage and computational load.

\emph{Efficient offloading mechanisms:} Extend the capabilities of models like FlexGen \cite{b16} and DeepSpeed Inference \cite{b5} by refining offloading techniques. This includes better utilization of CPU, GPU, and NVMe memory to handle larger models with fewer resources.

\emph{Resource-aware scheduling:} Implement intelligent scheduling mechanisms that consider the specific resource constraints of the hardware, optimizing the allocation of GPU, CPU, and memory resources for different types of tasks.

\subsection{Memory and Computation Optimization}
\label{memory_and_computation_optimization}

\emph{Advanced memory management:} Implement various techniques like dynamic catching, memory recycling, and efficient layer normalization (as presented in ByteTransformer \cite{b4} and LightSeq2 \cite{b52}) to overcome the memory overhead problem.

\emph{Mixed-precision Training} In order to significantly reduce training time and resource consumption without sacrificing accuracy, develop robust mixed-precision methods (like Megatron-LM \cite{b18} and LightSeq2 \cite{b52}).

\emph{Dynamic input handling:} Focusing on variable-length inputs, like ByteTransformer \cite{b4}, is seen as a promising area for improvement in ML, especially for NLP tasks that often deal with data of varying lengths. By developing more advanced algorithms to handle these inputs and minimize unnecessary computations, frameworks could achieve significant performance gains in NLP.

\subsection{Parallelism and Distribution}
\label{parallelism_and_distribution}

\emph{Adaptive parallelism:} Develop more advanced techniques that can dynamically adapt the parallelism strategy based on the model size and hardware configuration. This includes both data and model parallelism that can be adjusted on-the-fly to optimize performance.

\emph{Distributed training and inference:} Improve frameworks like PETALS \cite{b57} and CoLLiE \cite{lv2023collie} to better leverage distributed and heterogeneous hardware resources for efficient training and inference.

\subsection{Scalable and Modular Architecture}
\label{scalabile_and_modular_architecture}

\emph{Composable frameworks:} Design frameworks with modular components, similar to NLP-Fast \cite{b55}. These components act like building blocks for inference pipelines. Users can easily swap or optimize individual components independently, allowing for greater flexibility and customization.

\emph{Flexible APIs:} Create user-friendly APIs, like those in PETALS \cite{b57}. These APIs allow users to customize inference and fine-tuning processes according to their specific needs without having to make extensive changes to the underlying framework. This provides greater control and adaptability for different use cases.

\subsection{Performance Optimization Techniques}
\label{perfoamcen_optimization_techniques}

\emph{Adaptive algorithms:} Develop algorithms that can adapt to varying input sizes and sequences, optimizing both memory allocation and computational load dynamically.

\emph{Custom kernel implementations:} Continue to develop and refine custom kernel implementations for key operations like Softmax and LayerNorm to achieve better performance, as seen in TurboTransformers \cite{b11}. This could also involve hardware-specific optimizations for different GPU architectures.

\subsection{Advanced Compression and Quantization}
\label{advanced_compression_techniques}

\emph{Sophisticated compression techniques:} To reduce model size without significant accuracy loss instigate new methods for both lossless and lossy compression going beyond FlexGen's 4-bit quantization \cite{b16}.

\emph{Dynamic quantization:} Develop dynamic quantization techniques that adjust the precision of weights and activations in real time based on the computational requirements and available resources.

\section{Limitations}m
\label{limitaions}
In this section, we will present the limitations of our SLR.  Here, we acknowledge that while our review offers valuable insights, it is essential to consider its scope and boundaries. The limitations of  our SLR can be stated as follows:

\emph{Timeframe:} 
This SLR focused on studies published between 2017 and December 2023. While this timeframe deliberately captured a period of significant advancement in LLM optimization techniques, it is acknowledged that relevant research published before 2017 or after December 2023 might have been excluded. This could potentially limit the comprehensiveness of the analysis, particularly regarding foundational concepts or emerging advancements outside the chosen timeframe.

\emph{Search strategy:}
The chosen search queries might not have encompassed all possible relevant terminology used in LLM optimization research. This limitation could result in missing out on studies that use different terminologies or keywords to describe similar concepts and techniques.

\emph{Database coverage:} 
If the search excluded specific databases that are highly relevant to LLM research, significant studies might have been overlooked. Comprehensive database coverage is crucial to ensure the inclusion of all pertinent research.

\section*{Acknowledgment}
The authors are grateful to the members of the Applied Machine Learning Research Group of Óbuda University John von Neumann Faculty of Informatics for constructive comments and suggestions. The authors would also like to acknowledge the support of the Doctoral School of Applied Informatics and Applied Mathematics of Óbuda University.

\section*{List of Abbreviations}
\null
{\footnotesize
\begin{tabular}{rl}
AdaLomo & Low-Memory Optimization with Adaptive \\ & Learning Rate\\
BART & Bidirectional and Auto-Regressive Transformers\\
BERT & Bidirectional Encoder Representations from Transformers\\
BLOOM & BigScience Large Open-science Open-access Multilingual\\ & Language Model\\
CD &  Coordinate Descent \\
EET & Easy and Efficient Transformer\\
FPGA & Field Programmable Gate Arrays\\
FPTQ & Fine-Grained Post-Training Quantization \\
FT & Faster Transformer\\
GLM & General Language Model\\
GPT & Generative Pre-trained Transformer\\
GPU & Graphical Processing Unit\\
HAO & Hardware-Aware Optimization\\
IR & Information Retrieval\\
KV & Key Value\\
LAMBADA &  LAnguage Modeling Broadened to Account  \\ & for Discourse Aspects \\
LLaMA & Large Language Model Meta AI\\
LLM-QAT & LLM-Quantization-Aware Training \\
LM & Language Model\\
LOMO & Low-Memory Optimization \\
LoRA & Low-Rank Adaptation \\
MHA & Multi-Head Attention\\
MoE & Mixture-of-Experts\\
MMLU & Massive Multitask Language Understanding\\
NLP & Natural Language Processing\\
NN & Neural Network\\
OPT & Open Pre-trained Transformer\\
PET & Parameter Efficient Transformers\\
PetS & Parameter-Efficient Transformers Serving\\
PEFT & Parameter-Efficient Fine-Tuning \\
PIE & PET Inference Engine\\
PLM & Pre-trained Language Model\\
PRISMA & Preferred Reporting Items for Systematic Reviews\\ & and Meta-Analyses\\
PTM & Pre-Trained Model\\
PTQ & Post-Training Quantization \\ 
SLR & Systematic Literature Review\\
SWARM & Stochastically Wired Adaptively Rebalanced Model\\
VAE & Variational Autoencoder\\
W4A8 & 4-bit weights and 8-bit activations \\
\end{tabular}}

\bibliographystyle{ieeetr}
\bibliography{ref}

\begin{IEEEbiography}[{\includegraphics[width=1in,height=1.25in,clip,keepaspectratio]{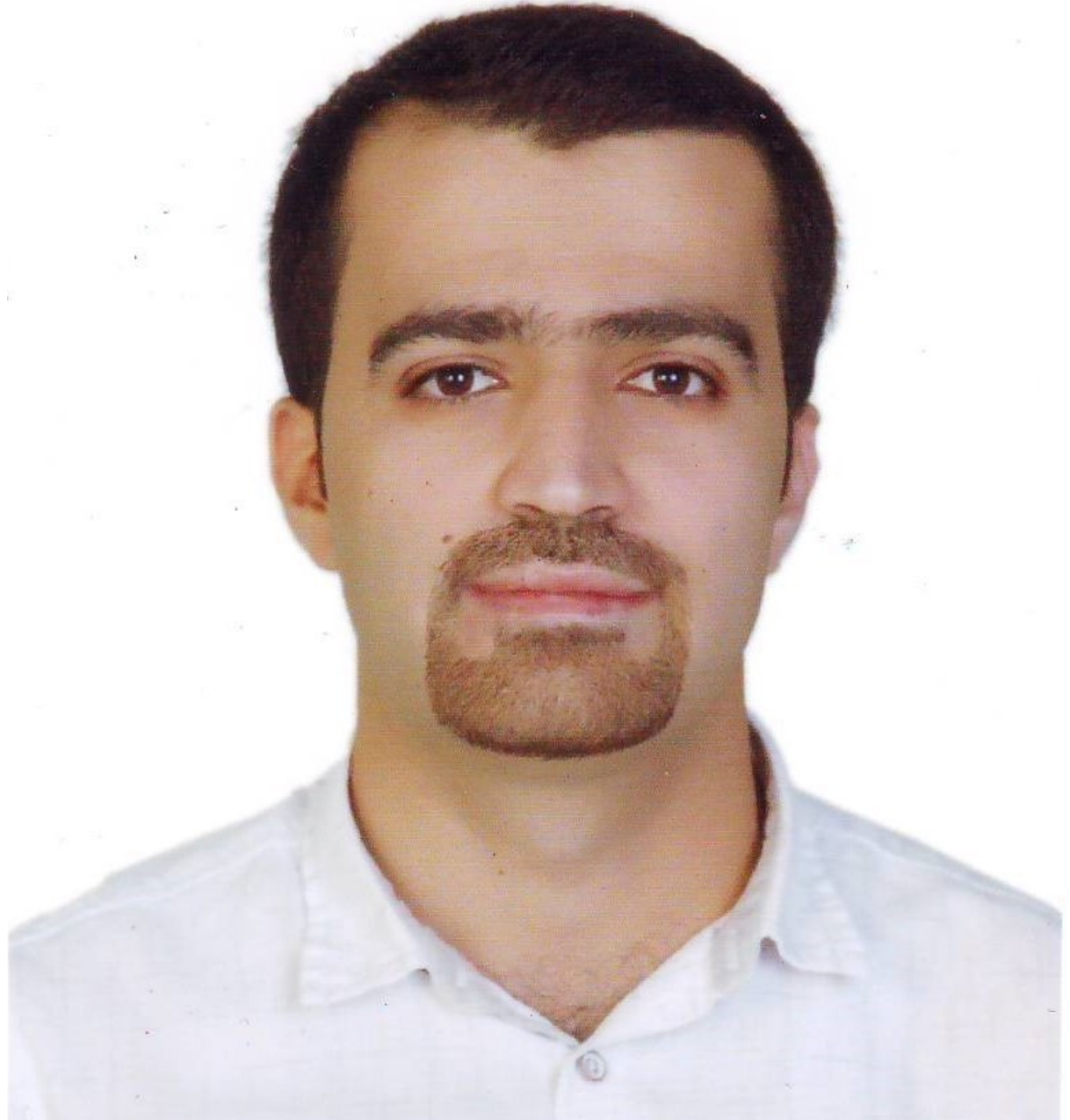}}]{Zhyar Rzgar K Rostam} (Member, IEEE) received the B.Sc. and M.Sc. degree in computer science from the University of Sulaimani, KRG-Iraq, in 2013 and 2019. He is currently pursuing a Ph.D. degree in Information Science and Technology at Óbuda University, Budapest, Hungary. His current research interests include large language models, scientific text classification, deep learning techniques, machine learning algorithms, and artificial intelligence.
\end{IEEEbiography}
\begin{IEEEbiography}[{\includegraphics[width=1in,height=1.25in,clip,keepaspectratio]{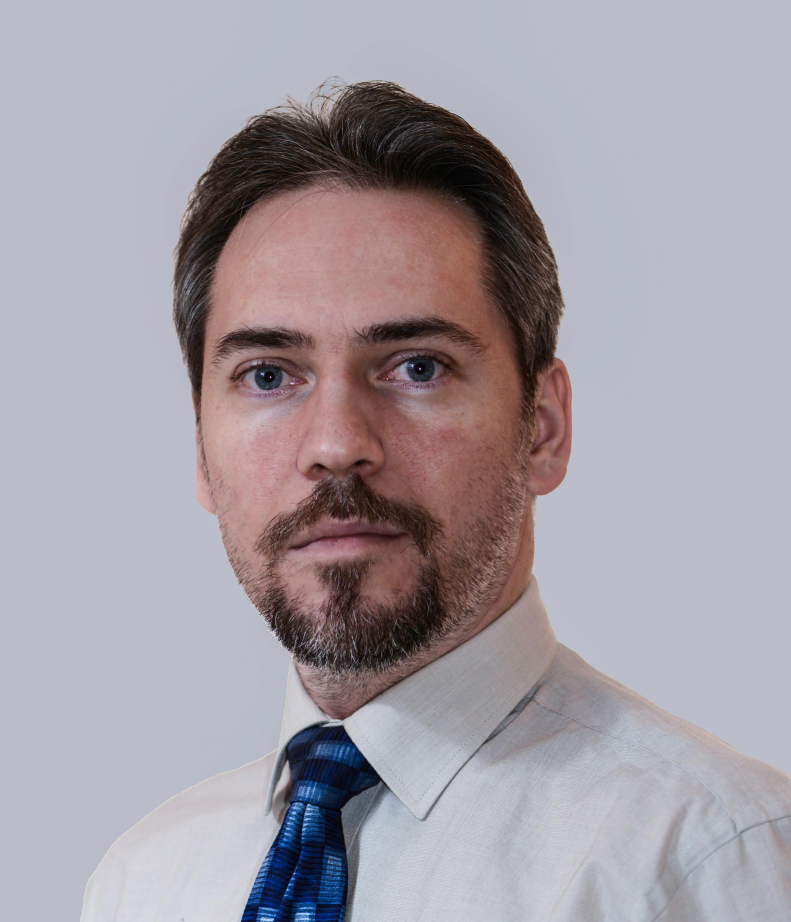}}]{SÁNDOR SZÉNÁSI} (Member, IEEE) received his PhD in 2013 from Doctoral School of Applied Informatics and Applied Mathematics of Óbuda University, Budapest, Hungary. Currently, he is a professor at the John von Neumann Faculty of Informatics, Óbuda University, Budapest, Hungary. His research areas are (data) parallel algorithms, GPU programming, and medical image processing. He engages both in theoretical fundamentals and in algorithmic issues with respect to realization of practical requirements and given constraints.
\end{IEEEbiography}
\begin{IEEEbiography}[{\includegraphics[width=1in,height=1.25in,clip,keepaspectratio]{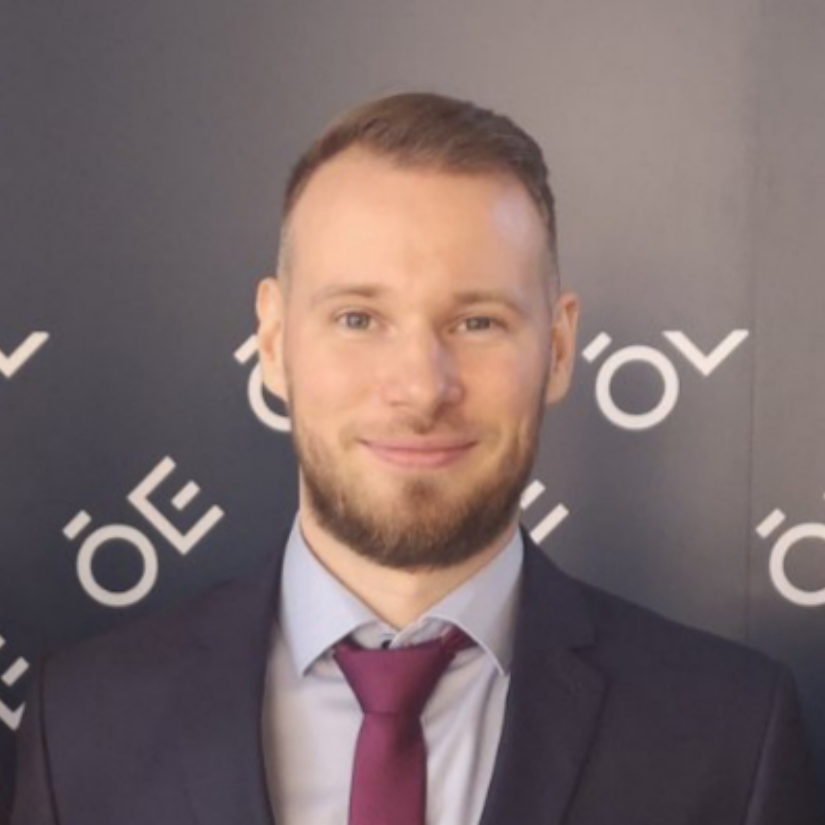}}]{Gábor Kertész} (senior member, IEEE) received his PhD in 2019 in Information Science and Technology; the main areas of his research was computer vision, parallel processing, deep machine learning. His current research interests include distributed deep learning, metric learning, and applied machine intelligence.

He is an associate professor and the vice-dean for research at Óbuda University John von Neumann Faculty of Informatics, Budapest, Hungary, also a part-time research-fellow at the HUN-REN SZTAKI (Institute for Computer Science and Control). He is the leader of the Applied Machine Learning Research Group at the John von Neumann Faculty of Informatics.

Dr.~Kertész is the founding president of the High Performance Computing division of the John von Neumann Computer Society, and the president of the IEEE Computational Intelligence Hungary Chapter.
\end{IEEEbiography}

\EOD

\end{document}